\PassOptionsToPackage{pagebackref}{hyperref}
\documentclass[10pt,a4paper]{paper}
\usepackage{fontawesome5}

\usepackage[authoryear, round]{natbib}

\usepackage{amsmath,amsfonts,bm}

\newcommand{\abs}[1]{\ensuremath{\left| #1 \right|}}

\def\eqref#1{equation~\ref{#1}}

\def\1{\bm{1}}

\DeclareMathAlphabet{\mathsfit}{\encodingdefault}{\sfdefault}{m}{sl}
\SetMathAlphabet{\mathsfit}{bold}{\encodingdefault}{\sfdefault}{bx}{n}

\DeclareMathOperator*{\argmin}{arg\,min}

\usepackage[table]{xcolor}
\usepackage[capitalise,noabbrev]{cleveref}
\usepackage{url}
\usepackage{xspace}
\usepackage{graphicx}
\usepackage{placeins}
\usepackage{amsmath}
\usepackage{amssymb}
\usepackage{algorithm}
\usepackage{algpseudocode}
\usepackage{subcaption}
\usepackage{enumitem}
\usepackage{booktabs}
\usepackage{float}
\usepackage{amsthm}
\usepackage{multirow}

\definecolor{mydarkgreen}{rgb}{0,0.45,0.08}
\definecolor{mydarkblue}{rgb}{0,0.1,0.6}
\definecolor{linkblue}{RGB}{0,70,140}

\definecolor{title_blue}{HTML}{204899}
\definecolor{cite_blue}{HTML}{044dc1}
\definecolor{cite_purple}{HTML}{7406a7}

\hypersetup{
    colorlinks = true,
    citecolor = {cite_blue},
    linkcolor = {cite_purple},
    urlcolor = {cite_purple},
}

\makeatletter
\renewcommand\footnoterule{%
  \kern 15\p@
  \hrule \@width 2in \kern 2.6\p@
  \vspace{4pt}
}
\makeatother

\newcommand{\ouralg}{\textsc{RTAQ}}

\newcommand{\eg}{\emph{e.g.}}

\title{RTAQ: Reasoning Termination-Aware Quantization for Mixture-of-Experts Architecture}

\reportnumber{}

\author[1]{Minjae Choi}
\author[2]{Seung-gyeom Kim}
\author[2]{Sukmin Yun}
\author[1,$\dagger$]{Insu Han}
\affil[1]{KAIST}
\affil[2]{Hanyang University}
\correspondingauthor{Contact: {\{minjaechoi, insu.han\}@kaist.ac.kr}, {\{skkim3533, sukminyun\}@hanyang.ac.kr}. ${}^\dagger$Corresponding author: Insu Han.}

\begin{abstract}
\textbf{Abstract:}
Extreme low-bit post-training quantization is key to serving large-scale mixture-of-experts (MoE) language models, yet reasoning models quantized to around two bits often collapse on long-horizon reasoning tasks. We show that a major source of this collapse is not a loss of knowledge but a failure to stop reasoning. When a near-2-bit reasoning model emits the end-of-thinking token (\texttt{</think>}), it almost always completes its answer, and for most methods the accuracy of such responses stays close to full precision; a large share of the remaining errors (up to 47\% for GPTQ) come from reasoning traces that never close. We trace this termination decision to a small subset of routed experts (approximately 2\% of all routed experts) that are consistently selected at the end-of-thinking position but are largely overlooked by standard sensitivity metrics. Building on this finding, we propose RTAQ (Reasoning-Termination-Aware Quantization), a weight-only quantization pipeline with three stages: (1) constructing a bank of activation-weighted, DP-optimal scalar codebooks for every expert and bit-width; (2) performing termination-aware bit allocation, which fixes termination experts at higher precision within the bit budget and assigns the remaining bits through knapsack optimization; and (3) applying grid-preserving recalibration, which reassigns codes using GPTQ-style error compensation and rescales expert outputs in under one second per expert. At an average bit-width of 2.0 for routed experts, RTAQ preserves over 94\textasciitilde 98\% of the full-precision performance on long-reasoning benchmarks such as AIME. On a single GPU, \ouralg{} also shrinks routed-expert memory by 6.3$\times$ and roughly doubles decoding throughput over BF16.

\vspace{0.15in}

\par\smallskip
\noindent\faGlobe\enspace\textbf{Project page:} \url{https://minjaechoics.github.io/rtaq-project/}
\par
\noindent\faGithub\enspace\textbf{Code:} \url{https://github.com/kaist-flexml-lab/RTAQ}
\end{abstract}

\begin{document}

\maketitle

\section{Introduction} \label{sec:intro}
Frontier language models have increasingly converged on the Mixture-of-Experts (MoE) Transformers, including GLM~\citep{zeng2026glm}, DeepSeek~\citep{liu2024deepseek} and MiniMax~\citep{chen2026minimax}. 
By activating only a few experts per token, MoE decouples model capacity from per-token compute, enabling models with hundreds of billions to trillions of parameters.
However, expert weights still dominate memory and must remain available during inference.
Indeed, more than 97\% of the parameters in Qwen3.6-35B model reside in MoE blocks.
Weight quantization is therefore particularly attractive for MoE models, and pushing routed-expert weights toward the 2-bit regime can substantially reduce their memory footprint.

However, aggressive quantization can cause a failure mode beyond accuracy degradation: reasoning models may fail to terminate properly.
Instead of emitting \texttt{</think>} and proceeding to the final answer, they may continue reasoning excessively or fall into repetitive loops~\citep{sui2025,lotfi2026quantized}.
This quantization-induced overthinking increases decoding latency and KV-cache usage in addition to degrading accuracy.
Existing mixed-precision methods allocate bits using generic criteria such as reconstruction error or loss sensitivity~\citep{dong2020hawqv2,shen2020q,hooper2025fgmp,chittyvenkata2025mopeq}, without explicitly accounting for reasoning termination.

Our key observation is that \emph{reasoning termination is concentrated in a small subset of routed experts}.
At positions immediately preceding \texttt{</think>}, \Cref{fig:te-hist} shows that most experts are rarely selected, whereas a small subset is activated far more frequently than expected under uniform routing.
We call these experts \emph{termination-aware experts}.
This observation suggests that preserving a very small, functionally important subset of experts at higher precision may protect reasoning termination without sacrificing the memory benefits of aggressive quantization.

\begin{figure}[t]
    \centering
    \vspace{-0.14in}
    \includegraphics[
        width=0.99\linewidth,
    ]{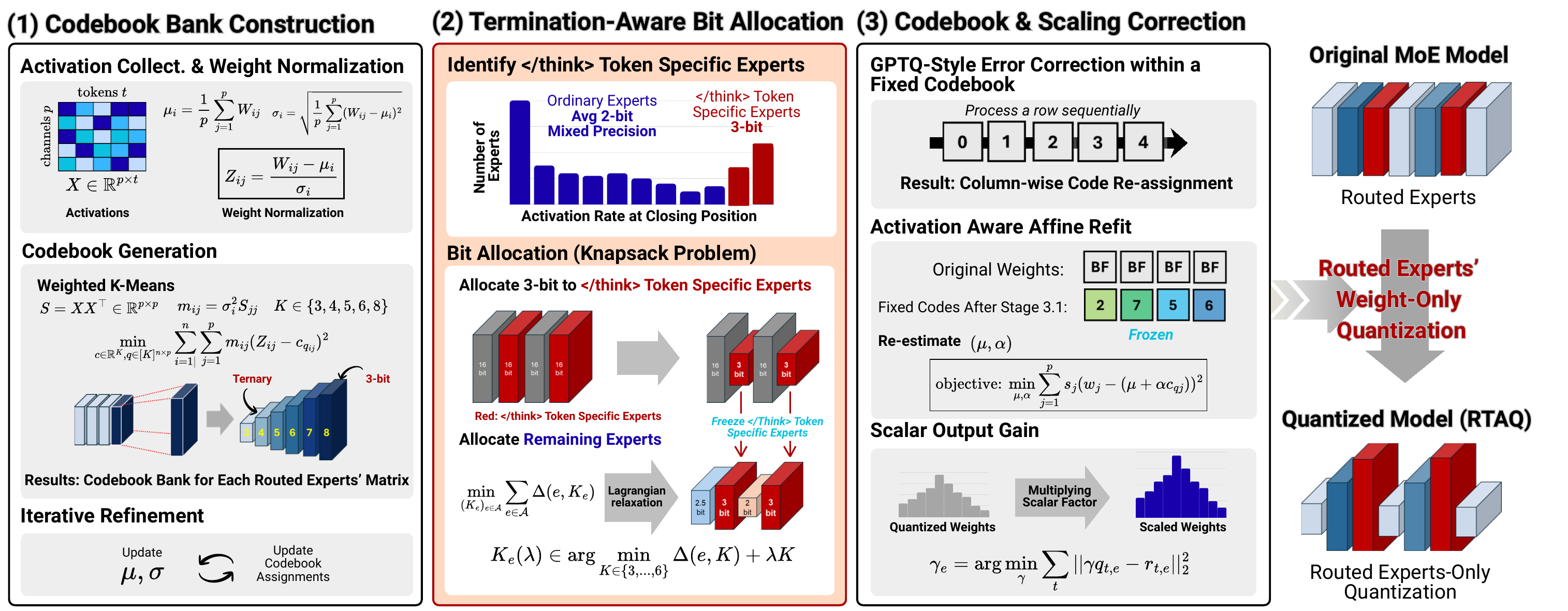}
    \caption{Overview of RTAQ (Reasoning Terminate-Aware Quantization).}
    \label{fig:overview}
\end{figure}

Motivated by this finding, we propose \ouralg{} (\textbf{R}easoning \textbf{T}ermination-\textbf{A}ware \textbf{Q}uantization), a mixed-precision quantization framework for near-2-bit MoE models.
\ouralg{} first constructs an activation-weighted codebook bank spanning multiple quantization levels for each routed expert.
It then identifies termination-aware experts from their activation rates at closing positions and preserves them at higher precision (3 bits).
For the remaining experts, \ouralg{} estimates loss-aware quantization sensitivity and allocates precision under a global bit budget through a Lagrangian relaxation of the resulting multiple-choice knapsack problem.
Finally, it refines the selected quantized experts using GPTQ-style error compensation and a scalar output correction, without changing the allocated codebook sizes or storage format. An overview of \ouralg{} is illustrated in \cref{fig:overview}.

We evaluate \ouralg{} on three sparse MoE models spanning 30B to 80B parameters and five reasoning and coding benchmarks.
At only 2.016--2.025 bits per routed-expert weight on average, \ouralg{} achieves the highest average accuracy among all evaluated quantization methods on every model, outperforming the strongest competing method.
Moreover, \ouralg{} keeps generation lengths close to those of the original BF16 models, demonstrating that preserving termination-aware experts helps mitigate the excessive reasoning induced by aggressive quantization.
Our main contributions are:
\begin{itemize}[leftmargin=1.5em]
    \item We identify a small subset of routed experts that are activated at unusually high rates immediately before \texttt{</think>}. We call them \emph{termination-aware experts} and use them to preserve reasoning termination under quantization.

    \item We introduce \ouralg{}, a termination-aware mixed-precision framework combining an activation-weighted codebook bank, high-precision protection of termination-aware experts, loss-aware sensitivity, Lagrangian bit allocation, and fixed-codebook GPTQ-style correction.

    \item Across three MoE models and five benchmarks, \ouralg{} achieves the highest accuracy at the lowest average bit-width among evaluated methods. With about 2.02 bits per routed-expert weight, it reaches $85.81\%$ average accuracy on Qwen3.6-35B-A3B (BF16: $87.85\%$) and retains 94--98\% of BF16 average accuracy across the three models, while keeping generation lengths near BF16.
\end{itemize}

\subsection{Related Work}

\textbf{Quantization of Mixture-of-Experts Models. }
Recent work reduces the memory footprint of Mixture-of-Experts (MoE) models through low-bit quantization.
Early methods~\citep{kim2023mixture, frantar2023qmoe} focused on extreme low-bit compression of expert weights (\eg, sub-1-bit, 2-bit), but uniform bit-widths ignore differences in quantization sensitivity~\citep{li2024examining}.
Subsequent works~\citep{huang2025mixture, duanmu2025mxmoe, Chitty-Venkata_2025_ICCV, chowdhury2026efficient} therefore introduce mixed-precision quantization, assigning bit-widths according to expert importance.
They use signals (\eg, Hessian-based sensitivity and reconstruction error) to preserve important experts at higher precision.
More recent work extends this idea to model-wide bit-width optimization across experts~\citep{deng2026gemq}.
Despite these advances, existing methods rely on generic quantization criteria or aggregate task performance, without explicitly considering each expert's role in reasoning termination.

\textbf{Low-Bit Quantization for Reasoning Models.}
Several studies have shown that low-bit quantization can substantially degrade LLM reasoning~\citep{liu2025quantization,zhang2026when}.
Quantized models may generate unnecessarily long reasoning traces~\citep{lian2026quantization}, exhibit overthinking and delayed commitment~\citep{lotfi2026quantized}, and show repetitive loops or termination failures at extreme low bit-widths~\citep{alimaskina2026extreme}.
Recent approaches leverage self-generated reasoning traces for calibration~\citep{wang2026attend} and reasoning-aware QAT to recover degraded reasoning capability~\citep{okoshi2026towards}.
These efforts mainly target model-wide quantization error reduction or the recovery of reasoning capability.
However, they do not explicitly consider which model components are critical to reasoning termination.
In contrast, we quantize a small subset of termination-aware experts to 3 bits and the remaining experts to an average of 2 bits.
\section{Preliminaries}
\label{sec:prelim}
We consider a Transformer in which every block contains a sparse Mixture-of-Experts (MoE) layer~\citep{shazeer2017outrageously,fedus2022switch} with a shared expert.
In each MoE layer, a router selects a small subset of experts for each token and combines their outputs using routing weights.

\paragraph{Quantization Representation.}
\label{sec:prelim_codebook}
We approximate a pretrained weight matrix $W\in\mathbb{R}^{n\times p}$ by
\begin{equation}
    \widehat{W}_{ij}=\mu_i+\sigma_i c_{q_{ij}},
    \qquad q_{ij}\in\{1,\ldots,K\},
    \label{eq:quant_representation}
\end{equation}
where $c\in\mathbb{R}^{K}$ contains the quantization levels shared across rows, $q_{ij}$ selects a codebook entry, and $\mu,\sigma\in\mathbb{R}^{n\times1}$ are row-wise offsets and scales.
We use $\log_2 K$ logical bits per weight to describe code precision; actual storage also depends on index packing and the storage of the codebook and row-wise affine parameters.

Given calibration inputs $X\in\mathbb{R}^{p\times t}$, layer-wise quantization minimizes the output reconstruction error~\citep{nagel2020adaround,frantar2023gptq}
\begin{equation}
    \begin{aligned}
    \mathcal{E}(\widehat{W})
    =\left\|(W-\widehat{W})X\right\|_F^2
    =\operatorname{tr}\!\left((W-\widehat{W})S(W-\widehat{W})^\top\right),
    \end{aligned}
    \label{eq:layerwise_objective}
\end{equation}
where $S=XX^\top$ is the unnormalized second-moment matrix.
Under a diagonal approximation to $S$, the objective reduces to $\sum_{i,j}S_{jj}(W_{ij}-\widehat{W}_{ij})^2$, weighting each squared error by the activation energy of its input dimension.
The off-diagonal entries capture interactions between input dimensions, which GPTQ-style error compensation uses to adjust the remaining weights after quantization.

\paragraph{Loss Sensitivity for Mixed-Precision Quantization.}
\label{sec:prelim_allocation}
Mixed-precision methods allocate precision under a budget according to the increase in task loss $\mathcal{L}$ caused by quantization~\citep{dong2019hawq,dong2020hawqv2}.
Here, $\mathcal{L}$ is the model's average next-token prediction loss, so sensitivity measures the effect on the model's predictions beyond local reconstruction error.
If quantizing a block changes its vectorized output $h$ to $h+\delta$, a second-order Taylor expansion gives
\begin{equation}
    \Delta\mathcal{L}
    \approx g^{\top}\delta+\frac{1}{2}\delta^{\top}H\delta,
    \label{eq:second_order_sensitivity}
\end{equation}
where $g=\nabla_h\mathcal{L}$ and $H=\nabla_h^2\mathcal{L}$ are evaluated at the original BF16 model.
The quadratic term is commonly used as a sensitivity surrogate~\citep{dong2019hawq}. In practice, the curvature can be approximated using the empirical Fisher matrix or its diagonal approximation~\citep{martens2020new,kunstner2019limitations}. We adopt the diagonal approximation to obtain computationally tractable expert-wise sensitivity estimates.

\paragraph{Reasoning Termination.}
\label{sec:prelim_termination}
Recent language models that support \emph{thinking} mode~\citep{openai2024,guo2025deepseek,yang2025} can generate an explicit chain of thought before producing a final answer.
For example, Qwen3 models~\citep{yang2025qwen3} enclose the reasoning trace between \texttt{<think>} and \texttt{</think>}, then generate the final answer followed by an end-of-sequence token.
A response \emph{terminates} if it emits the end-of-sequence token within the generation budget and is \emph{truncated} otherwise.

Low-bit quantization can degrade reasoning performance~\citep{liu2025quantization}.
At extreme bit-widths, we focus on \emph{reasoning non-termination}, in which the reasoning trace loops or drifts until the generation budget is exhausted without emitting \texttt{</think>} or producing a final answer.
This failure is related to \emph{overthinking} observed in full-precision reasoning models, where reasoning continues beyond what the problem requires even when the response eventually terminates~\citep{chen2024overthinking,sui2025}.

\section{\ouralg: Reasoning Termination-Aware Quantization for MoE}
\label{sec:method}
\label{sec:main}

We introduce \ouralg, a mixed-precision quantization framework for routed experts that preserves higher precision for termination-aware experts.
First, we construct a \emph{codebook bank} containing multiple quantization levels for each expert.
Second, we allocate adaptive bit-widths across experts under a target budget while protecting termination-aware experts.
Finally, we refine the selected quantized experts without changing their codebooks.

\subsection{Stage I: Activation-Weighted Codebook Bank}
\label{sec:codebook_bank}

We first construct a {\it codebook bank} for each routed-expert weight, covering multiple quantization levels (i.e., codebook size) $K\in\{3,\dots,8\}$. 
The bank is constructed once offline and subsequently reused for different target bit budgets.
Given a matrix $W \in \mathbb{R}^{n \times p}$, we first normalize each row as
\begin{equation}
Z_{ij} = \frac{W_{ij}-\mu_i}{\sigma_i},\qquad \mu_i = \frac{1}{p}\sum_{j=1}^{p}W_{ij},\qquad
\sigma_i = \sqrt{\frac{1}{p}\sum_{j=1}^{p}\big(W_{ij}-\mu_i\big)^2},\label{eq:normalization}
\end{equation}
where $\mu, \sigma \in \mathbb{R}^{n \times 1}$ are the row-wise means and the standard deviations, respectively.
These vectors incur only a small quantization overhead, 
as each $(\mu_i,\sigma_i)$ is shared across all entries in row $i$.

For each quantization level $K \in \{3, \dots, 8\}$, we obtain a codebook $c \in \mathbb{R}^{K}$ by applying weighted $K$-means clustering to vectorized $Z$.
Rather than weighting all coordinates equally, we utilize their expected contribution to the layer output through an \emph{activation-weighted mass}.
Formally, let $X \in \mathbb{R}^{p \times t}$ denote calibration activations %
and denote $S = XX^\top \in\mathbb{R}^{p \times p}$. For each normalized weight $Z_{ij}$ for $i \in [n], j \in [p]$, we assign the mass $m_{ij} = \sigma_i^2 S_{jj}$ where 
$S_{jj}$ measures the activation energy associated with input dimension $j$.
The codebook and assignments are obtained by minimizing
\begin{equation}
\min_{c \in \mathbb{R}^K, q \in [K]^{n \times p}}
\sum_{i=1}^n \sum_{j=1}^p
m_{ij}
\left(
Z_{ij}-c_{q_{ij}}
\right)^2.
\end{equation}

We then refine the affine parameters $\mu$ and $\sigma$ to further reduce the reconstruction error.
Starting from the row-wise statistics in \cref{eq:normalization} and the weighted $K$-means solution, we perform alternating optimization: with $\mu$ fixed, we optimize $\sigma$; with $\sigma$ fixed, we optimize $\mu$.
After each update, we recompute the normalized weights and update their codeword assignments accordingly.
We repeat these steps 5 times.
After this offline stage, each routed expert is equipped with a codebook bank spanning $\log_2 3$ to $3$ bits per weight, from which a codebook can later be selected according to the desired expert-wise bit allocation.
Additional details are provided in \cref{app:reuse_bank}.

\subsection{Stage II: Termination-Aware Mixed-Precision Allocation}
\label{sec:bit_allocation}

Heavily quantized models often fail to terminate their reasoning at the appropriate time, resulting in excessive reasoning (overthinking) or repetitive loops~\citep{sui2025,lotfi2026quantized}.
We find that in MoE models, the prediction of the end-of-reasoning token \texttt{</think>} is routed to a small and consistent subset of experts, which we call \emph{termination-aware experts}.
We first identify these experts via empirical activation rates and protect them from aggressive quantization and then allocate precision to the remaining experts under a global bit budget.
\begin{figure}[t]
\centering
\captionsetup[subfigure]{skip=0pt}
\hspace{-0.2in}
\begin{subfigure}[b]{0.49\linewidth}
    \centering
    \includegraphics[width=\linewidth]{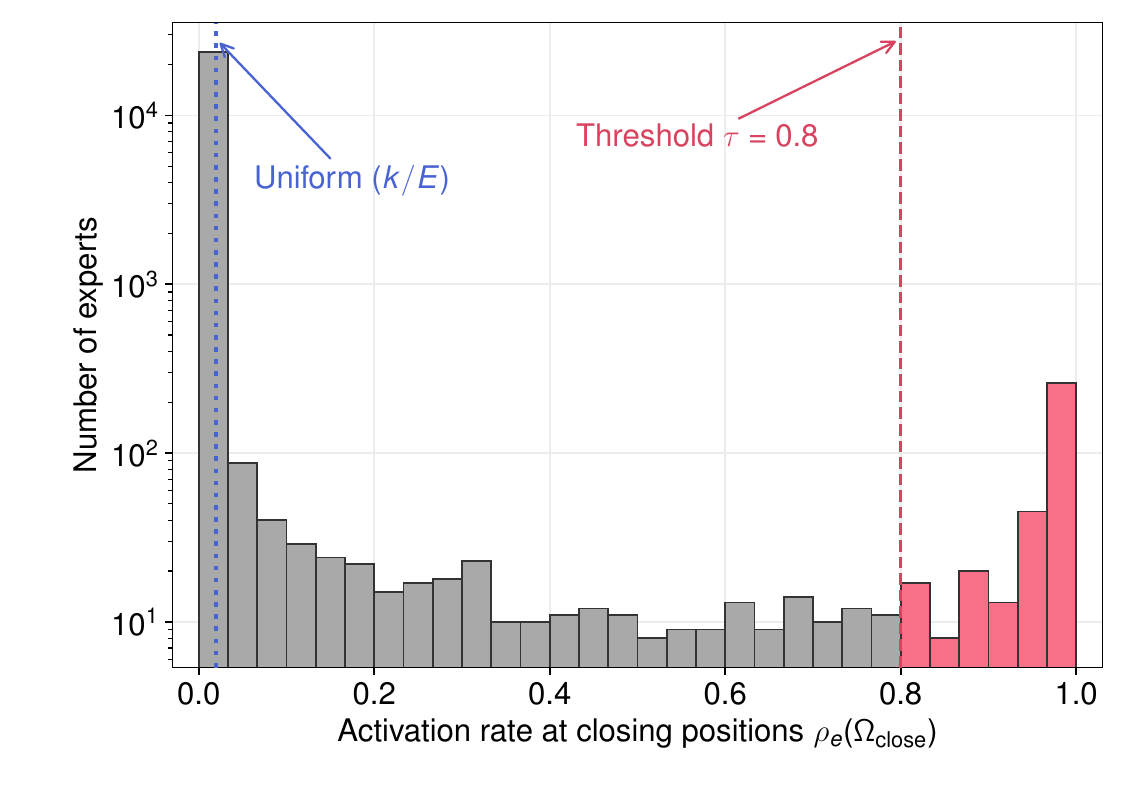}
    \caption{}
    \label{fig:te-hist}
\end{subfigure}
% \hfill
\begin{subfigure}[b]{0.49\linewidth}
    \centering
    \includegraphics[width=\linewidth]{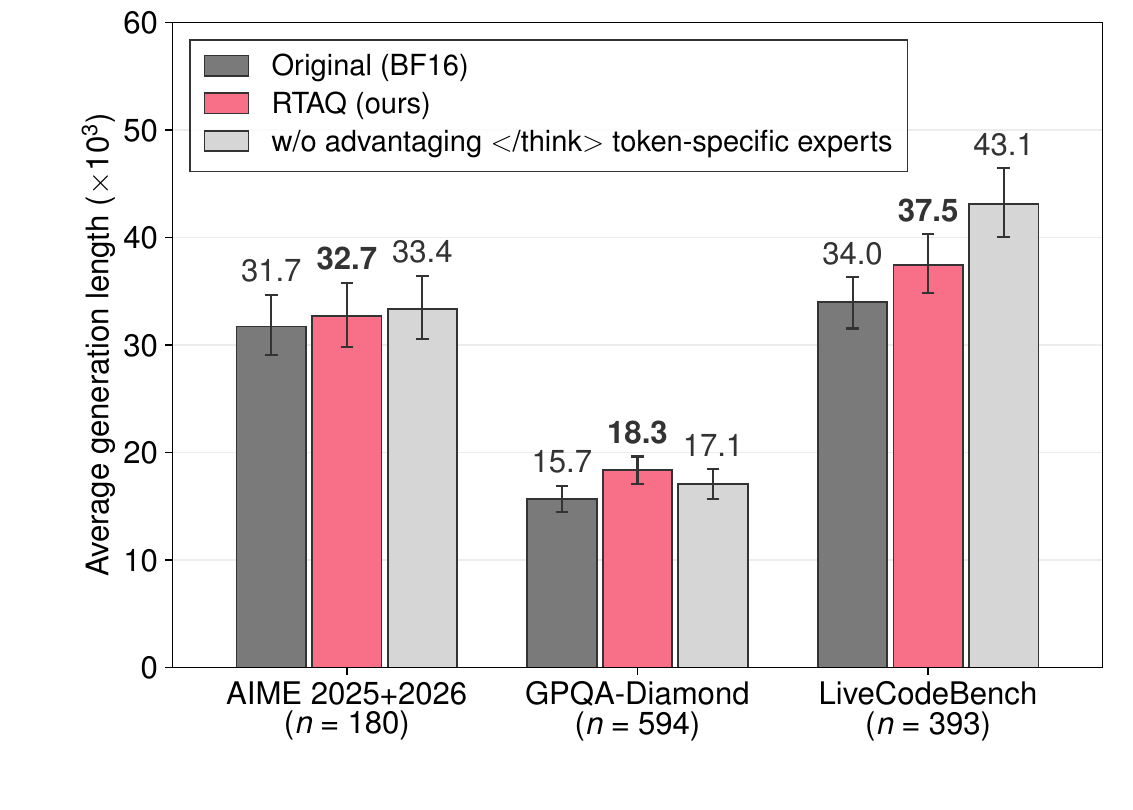}
    \caption{
    }
    \label{fig:te-removal}
\end{subfigure}
\vspace{-0.05in}
\caption{
\textbf{Emergence of termination-aware experts.}
(a) Distribution of routed-expert activation rates at
closing positions for Qwen3-Next-80B-A3B on GPQA-Diamond.
(b) Average generation length with and without
termination-aware expert pinning.
The effect of pinning is benchmark-dependent:
it reduces generation length on tasks where
reasoning non-termination is prevalent,
while the effect is less pronounced or reversed
on tasks where most responses already terminate.
}
\label{fig:te-existence}
\end{figure}

\paragraph{Reasoning Termination-Aware Experts.}\label{sec:termination_experts}
Consider a MoE model with $L$ layers and $E$ routed experts per layer, where each router selects $k$ experts for every token.
Let $\mathcal{E}$ denote the set of all routed experts across all layers, $|\mathcal{E}| = LE$. 
We collect $N$ sequences generated by the model, $\{x^{(i)}\}_{i=1}^{N}$ where $x^{(i)} = (x^{(i)}_1, \dots, x^{(i)}_{T_i})$ consists of the $i$-th prompt followed by the generated response.
We keep only sequences that contain the closing token \texttt{</think>}. 
Let $\mathcal{T}_{i,t} \subseteq \mathcal{E}$ denote the set of experts activated at position $t$ of $x^{(i)}$, collected over all layers. 

We call position $t$ a \emph{closing position} if the next token is \texttt{</think>}, i.e., $x^{(i)}_{t+1}=\texttt{</think>}$.
Thus, the closing position represents the state at which the model predicts the termination of its reasoning process.
Let $c_i\in[T_i-1]$ denote the closing position of sequence $i$, and define
\begin{equation}
    \Omega_{\mathrm{close}} = \{ (i, c_i) : i \in [N] \}.
\end{equation}

For any set of token positions $\Omega$, we define the \emph{activation rate} of expert $e\in\mathcal{E}$ as
\begin{equation}
    \rho_{e}(\Omega)
    =
    \frac{1}{|\Omega|}
    \sum_{(i,t) \in \Omega}
    \mathbb{I}\!\left[ e \in \mathcal{T}_{i,t} \right]
    \in [0,1],
    \label{eq:selection_rate}
\end{equation}
where $\mathbb{I}[\cdot]$ denotes the indicator function.
Given a threshold $\tau\in(0,1)$, we define the set of \emph{termination-aware experts} as
\begin{equation}
    \mathcal{P} :=
    \left\{ e \in \mathcal{E} \;:\;
    \rho_{e}(\Omega_{\mathrm{close}}) \ge \tau
    \right\}.
    \label{eq:termination_aware}
\end{equation}
That is, an expert is termination-aware if it is activated on at least a fraction $\tau$ of the closing positions.

\Cref{fig:te-existence} demonstrates the empirical evidence of termination-aware experts and statistics. 
\Cref{fig:te-hist} shows that activation rates at closing positions are highly skewed across routed experts. Most experts are rarely selected, while a small subset is activated far more frequently than expected under uniform routing, where the expected rate is $k/E$. This motivates preserving these \emph{termination-aware experts} during quantization. In practice, we set $\tau=0.8$.
\Cref{fig:te-removal} compares RTAQ with a variant that drops the 3-bit protection of $\mathcal{P}$ under the same bit budget. On LiveCodeBench, where non-termination is most frequent, dropping the protection lengthens generations from 37.5K to 43.1K tokens. On GPQA-Diamond, where about 95\% of responses already terminate, the unprotected variant is slightly shorter (17.1K vs.\ 18.3K) but less accurate when it terminates (78.9\% vs.\ 81.9\%). Protecting $\mathcal{P}$ thus matters most where termination is the bottleneck.

We therefore assign these experts in $\mathcal{P}$ to a higher bit (i.e., $K=8$ levels) and allocate the remaining bit budget adaptively across the other experts in $\mathcal{E} \setminus\mathcal{P}$.

\paragraph{Loss-Aware Expert Sensitivity.}\label{sec:bf16_cost}

To allocate adaptive bit-widths across experts, we quantify the sensitivity of expert $e$ to $K$-level quantization through the loss-aware perturbation of its MoE block.
Second-order sensitivity measures of this form have been widely used to guide mixed-precision quantization~\citep{dong2020hawqv2,shen2020q,hooper2025fgmp,chittyvenkata2025mopeq,autoquantize}.
Given a calibration sequence $x=(x_1,\ldots,x_T)$, let
$
\mathcal{L}(x)
=
\frac{1}{T}\sum_{t=1}^{T}\ell_t
$
denote the average next-token prediction loss, and let $h_t$ be the output of the MoE block containing expert $e$ at token position $t$.
We define
\begin{equation}
g_t
=
\frac{\partial \mathcal{L}(x)}{\partial h_t},
\qquad
\delta_t(e,K)
=
w_{t,e}
\left(
f_{e,K}(u_t)-f_e(u_t)
\right),
\label{eq1}
\end{equation}
where $u_t$ is the input to the MoE layer, $w_{t,e}$ is the routing weight assigned to expert $e$, and $f_e$ and $f_{e,K}$ denote the original and $K$-level quantized expert functions, respectively.
We set $w_{t,e}=0$ when expert $e$ is not selected.
Thus, $\delta_t(e,K)$ represents the local perturbation to the MoE output induced by quantizing only expert $e$.
Using the coordinate-wise diagonal of the empirical Fisher as the loss curvature, we define the sensitivity of the $(e,K)$ pair on sequence $x$ as
\begin{equation}
\begin{aligned}
\Delta(e,K;x)
&=
\frac{T}{2}
\sum_{t=1}^{T}
\delta_t(e,K)^\top
\operatorname{diag}(g_t \odot g_t)
\delta_t(e,K)=
\frac{T}{2}
\sum_{t=1}^{T}
\sum_{d=1}^{D}
g_{t,d}^{2}
\delta_{t,d}(e,K)^{2}.
\end{aligned}
\label{eq:sens}
\end{equation}
We approximate the curvature using the coordinate-wise diagonal of the empirical Fisher matrix, neglecting interactions between output dimensions.
The factor $T$ accounts for the normalization of the sequence-averaged loss $\mathcal{L}(x)$.
Thus, each token-wise curvature block is approximated as
$H_t \approx T\operatorname{diag}(g_t \odot g_t)$.

Intuitively, $\Delta(e,K;x)$ is large when quantizing expert $e$ induces a perturbation along directions to which the prediction loss is highly sensitive, while ignoring tokens that are not routed to expert $e$, i.e., $w_{t,e}=0$.
Given a calibration set $\{x^{(i)}\}_{i=1}^{N}$, we estimate the sensitivity of each expert--level pair by averaging across sequences: $\Delta(e,K)=\frac{1}{N} \sum_{i=1}^{N} \Delta\left(e,K;x^{(i)}\right)$.
Additional details on the curvature approximation are provided in \cref{sec:prelim_allocation}.

\paragraph{Mixed-Precision Allocation via Lagrangian Relaxation.}
Let $\mathcal{A}=\mathcal{E}\setminus\mathcal{P}$ denote the set of non-termination-aware experts.
Given a target average bit-width $b_{\mathrm{target}}$, we reserve the maximum codebook size
$K_{\max}=8$ for termination-aware experts in $\mathcal{P}$.
The remaining bit budget for experts in $\mathcal{A}$ is therefore $B = |\mathcal{E}| b_{\mathrm{target}} - |\mathcal{P}| \log_2 K_{\max}$.
The mixed-precision allocation problem is then formulated as
\begin{equation}
    \min_{(K_e)_{e\in\mathcal{A}}}
    \sum_{e\in\mathcal{A}} \Delta(e,K_e)
    \qquad
    \text{s.t.}
    \qquad
    \sum_{e\in\mathcal{A}} \log_2 K_e \le B,
    \quad
    K_e\in\{3,\dots,6\}.
    \label{eq:mckp}
\end{equation}
This problem is a multiple-choice knapsack problem \citep{sinha1979multiple} which is NP-hard in general. 
A na\"ive approach is 
to greedily increase the precision of the expert that yields the largest sensitivity reduction per additional bit.
However, it considers only local transitions between neighboring configurations and can be suboptimal when the sensitivity--precision tradeoff is non-convex.
We instead use a Lagrangian relaxation that compares all candidate precisions for each expert simultaneously~\citep{everett1963generalized,shoham1988efficient}.
Full details are provided in \cref{app:reuse_allocation}.

\subsection{Stage III: Codebook and Scaling Correction}
\label{sec:codebook_gptq}

After mixed-precision allocation, we further refine each selected quantized expert without changing its quantization level, codebook, or storage format.
Specifically, we improve the discrete codeword assignments using GPTQ-style error compensation, update the row-wise affine parameters and apply a scalar output correction to compensate for residual expert-level distortion.
Importantly, the normalized codebook selected in the previous stage remains fixed throughout this stage.

\paragraph{GPTQ-Style Error Correction within a Fixed Codebook.}
\label{sec:gptq-correction}

We first refine the codeword assignments using an OBS/GPTQ-style sequential compensation procedure.
Unlike standard GPTQ, which quantizes weights onto a predefined uniform grid, we quantize each compensated weight to its nearest entry in the selected codebook.
For a current compensated weight value $w_j$, row-wise affine parameters $(\mu,\sigma)$, and a fixed normalized codebook $c\in\mathbb{R}^{K}$, we assign
\begin{equation}
q_j
=
\argmin_{q\in[K]}
\left|
w_j-\left(\mu+\sigma c_q\right)
\right|,
\label{eq:fixed_code_assignment}
\end{equation}
and reconstruct it as 
$
\widehat{w}_j
=
\mu+\sigma c_{q_j}$,
$\varepsilon_j = w_j-\widehat{w}_j$.
As in GPTQ, the quantization error $\varepsilon_j$ is propagated to the remaining unprocessed weights using an inverse-curvature-based compensation update.
Crucially, \cref{eq:fixed_code_assignment} is applied to the \emph{current compensated value} of $w_j$, rather than its original value, so  each assignment accounts for errors introduced by previously quantized weights.

This procedure does not construct a new codebook or alter the allocated quantization level. It only updates the discrete assignments within the fixed normalized codebook. Therefore, the corrected expert retains the same codebook size and storage format, while benefiting from GPTQ-style error compensation.

\paragraph{Scalar Output Gain.}
Even after weight-level correction, residual error can remain in the expert output.
In particular, quantization errors in the gate and up projections pass through the nonlinear SiLU gating and subsequently interact with the quantized down projection, which can introduce a systematic bias in the output magnitude.
We correct this effect using a single scalar gain per expert.

For a calibration token $t$, let $q_{t,e} = w_{t,e} f_{e,K}(x_t)$, $r_{t,e} = w_{t,e} f_e(x_t)$
where $f_{e,K}$ and $f_e$ denote the quantized expert and the original BF16 expert, respectively, and $w_{t,e}$ is the routing weight.
We choose the gain $\gamma_e$ that best aligns the quantized expert output with its BF16 counterpart:
\begin{align}
\gamma_e
=
\argmin_{\gamma \in \mathbb{R}}
\sum_t
\abs{\gamma q_{t,e} - r_{t,e} }_2^2
=
\frac{
\sum_t
\left\langle q_{t,e},r_{t,e}\right\rangle
}{
\sum_t
|q_{t,e}|_2^2
}. \label{eq:scalar_gain}
\end{align}
This scalar corrects systematic output shrinkage or amplification that is not captured by local weight reconstruction alone.
We fold $\gamma_e$ into the affine scale and offset of the expert's down projection, so it introduces no additional inference operation and requires no additional per-weight storage.
Importantly, neither the discrete codes nor the codebook are modified.
Further details are provided in \cref{sec:damping_selection}.

\section{Experiments}
\label{sec:exp}

\begin{table*}[tb]
\caption{
Performance comparison of various quantization methods across 3 models on math, reasoning and coding benchmarks.
All reported results are accuracies (\%; higher is better). ``Bits'' denotes the average number of bits per weight in the routed experts. Baseline denotes the original unquantized model in BF16 precision. Bold indicates the best result for each model and benchmark.
}
\vspace{-0.2in}
\label{tab:acc-only}
\begin{center}
\small
\setlength{\tabcolsep}{3.5pt}
%\resizebox{\textwidth}{!}{
% \begin{tabular}{lcccccccc}
\begin{tabularx}{\textwidth}{l*{8}{>{\centering\arraybackslash}X}}
\toprule
\multirow{2}*{Method} & \multirow{2}*{Bits}
& \multirow{2}*{\bf GSM8K}
& \multirow{2}*{\bf AIME25}
& \multirow{2}*{\bf AIME26} 
& \multicolumn{2}{c}{\bf GPQA-Diamond}
& \multirow{2}*{\bf LCB} 
& \multirow{2}*{\bf Average} \\
& & & & & Strict. & Flex. & & \\
\midrule
\rowcolor{gray!20}\multicolumn{9}{c}{\textbf{Qwen3.6-35B-A3B}} \\
\midrule
Baseline & 16 & 96.8 & 91.1 & 91.1 & 82.3 & 83.0 & 77.6 & 87.85 \\
\midrule[0.1pt]
GPTQ~{\small \citep{frantar2022gptq}} & 2.250 & 95.7 & 80.0 & 81.1 & 71.2 & 72.9 & 50.1 & 75.79 \\
MC-MoE~{\small \citep{huang2025mixture}}& 2.358 & 96.3 & 88.9 & 85.6 & 71.9 & 72.6 & {\bf 72.3} & 83.07 \\
GSQ~{\small \citep{dadgarnia2026gsq}} & 2.125 & 96.0 & 84.4 & 86.7 & {\bf 79.8} & {\bf 80.6} & 66.4 & 82.74 \\
TileQ~{\small \citep{gu2026tileq}} & 2.320 & 95.2 & 82.2 & 86.7 & 73.7 & 71.7 & 49.4 & 77.24\\
\ouralg \xspace(Ours) & 2.022 & {\bf 96.6} & {\bf 90.0} & {\bf 93.3} & 78.6 & 79.3 & 70.2 & {\bf 85.81} \\

\midrule
\rowcolor{gray!20}\multicolumn{9}{c}{\textbf{Nemotron-3-Nano-30B-A3B}} \\
\midrule
Baseline & 16 & 96.1 & 87.8 & 90.0 & 72.7 & 73.1 & 66.2 & 82.60 \\
\midrule[0.1pt]
GPTQ~{\small \citep{frantar2022gptq}} & 2.250 & {\bf 94.7} & {\bf 85.6} & 80.0 & 58.9 & 59.4 & 51.9 & 74.27 \\
MC-MoE~{\small \citep{huang2025mixture}} & 2.358 & 94.2 & 81.1 & 84.4 & 63.8 & 63.8 & 58.0 & 76.30 \\
GSQ~{\small \citep{dadgarnia2026gsq}} & 2.125 & 93.6 & 82.2 & 82.2 & 38.2 & 37.7 & 60.1 & 71.21 \\
TileQ~{\small \citep{gu2026tileq}} & 2.301 & 94.0 & 81.1 & 74.4 & 60.6 & 60.4 & 52.2 & 72.44\\
\ouralg \xspace(Ours) & 2.025 & 94.3 & 81.1 & {\bf 85.6} & {\bf 67.2} & {\bf 67.0} & {\bf 60.3} & {\bf 77.68} \\

\midrule
\rowcolor{gray!20}\multicolumn{9}{c}{\textbf{Qwen3-Next-80B-A3B-Thinking}} \\
\midrule
Baseline & 16 & 96.6 & 85.6 & 88.9 & 75.6 & 75.3 & 68.2 & 82.95 \\
\midrule[0.1pt]
GPTQ~{\small \citep{frantar2022gptq}} & 2.250 & 96.1 & 78.9 & 87.8 & 65.2 & 63.3 & 58.3 & 77.07 \\
MC-MoE~{\small \citep{huang2025mixture}} & 2.358 & 96.1 & {\bf 84.4} & {\bf 91.1} & 59.6 & 58.8 & 57.8 & 77.72 \\
GSQ~{\small \citep{dadgarnia2026gsq}} & 2.125 & {\bf 96.4} & 83.3 & 84.4 & 62.3 & 58.2 & 56.0 & 76.07 \\
TileQ~{\small \citep{gu2026tileq}} & 2.307 & 96.1 & 80.0 & 87.8 & 68.2 & 68.7 & 61.6 & 78.79\\
\ouralg \xspace(Ours) & 2.016 & {\bf 96.4} & {\bf 84.4} & 88.9 & {\bf 72.2} & {\bf 70.5} & {\bf 64.1} & {\bf 81.03} \\

\bottomrule
\end{tabularx}
%}
\end{center}
\end{table*}

\paragraph{Setup.}
We compare \ouralg{} against various post-training quantization baselines including 
GPTQ~\citep{frantar2022gptq} as a standard weight-only quantization method, 
MC-MOE~\citep{huang2025mixture} and TileQ~\citep{gu2026tileq}
as MoE-specific quantization methods, and  
GSQ~\citep{dadgarnia2026gsq}, a recent low-bit scalar quantization method.
We evaluate these methods on three models spanning 30B to 80B parameters: Qwen3.6-35B-A3B, Nemotron-3-Nano-30B-A3B and Qwen3-Next-80B-A3B-Thinking.
All models adopt sparse MoE architectures, with more than 97\% of their parameters residing in the routed experts. 
We therefore quantize only the routed-expert parameters and choose method-specific hyperparameters such that each method uses approximately 2 bits per routed expert parameter on average.

We evaluate the resulting quantized models on five benchmarks: GSM8K~\citep{cobbe2021training}, AIME25/26~\citep{aime25,aime26}, GPQA-Diamond~\citep{rein2023gpqa} and LiveCodeBench~\citep{jain2025livecodebench}, spanning mathematical reasoning, scientific reasoning, and code generation. For GPQA-Diamond, we report results under two evaluation criteria, strict-match and flexible-extract.
All experiments were conducted on NVIDIA RTX PRO 6000 GPUs (with 96 GB VRAM), using two GPUs for 30B-parameter models and four GPUs for the 80B-parameter model.

\subsection{Accuracy Preservation under Near-2-Bit Quantization}

\Cref{tab:acc-only} summarizes accuracy on five benchmarks across three models under a near-2-bit routed-expert budget. Averaged over the benchmarks, \ouralg{} achieves the highest accuracy among all quantized methods on every model while using the fewest bits per routed-expert weight.Compared with the strongest baseline on each model (MC-MoE on Qwen3.6-35B-A3B and Nemotron-3-Nano-30B-A3B, TileQ on Qwen3-Next-80B-A3B-Thinking), \ouralg{} improves average accuracy by 2.74, 1.38, and 2.24 points, respectively. It also remains close to BF16, retaining 97.7\% of the original average accuracy on both Qwen models.
Moreover, we found that \ouralg{} is substantially more robust across individual model--benchmark pairs. While competing methods suffer large drops on challenging tasks such as GPQA-Diamond and LiveCodeBench, the largest BF16-relative degradation of \ouralg{} is only 7.4 points, compared with 16.5--35.4 points for the baselines. These results show that \ouralg{} preserves accuracy more consistently in the near-2-bit regime.

\subsection{Accuracy--Generation Length Trade-off}

\begin{figure}[t]
    \centering
    \includegraphics[width=\linewidth]{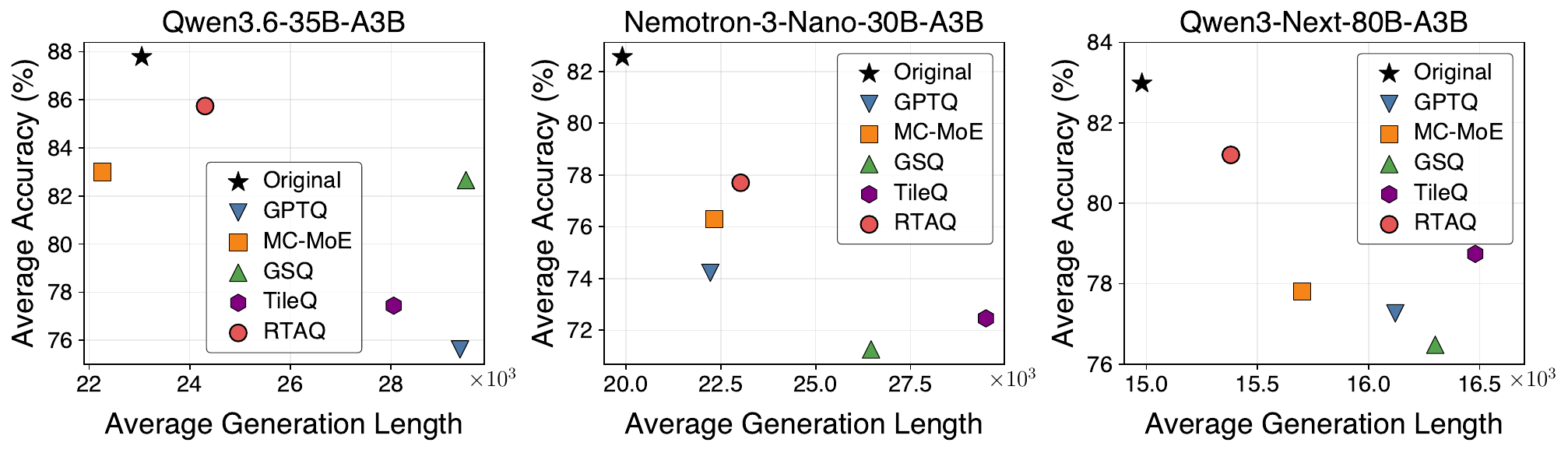}
    \caption{Accuracy–length trade-off across three models. Each point reports the average accuracy and generation length across five benchmarks. The original BF16 model (black, star) is included as a reference. Higher accuracy and shorter generation length are preferred (upper left). RTAQ (red, circle) consistently provides a favorable trade-off, substantially preserving the accuracy of the original model while maintaining generation lengths close to the original.}
    \label{fig:pareto}
\end{figure}

Low-bit quantization can also alter the reasoning behavior of MoE models, often causing excessively long reasoning traces or failure to emit \texttt{</think>} at an appropriate point. \Cref{fig:pareto} summarizes this quantization-induced overthinking through the accuracy--generation length trade-off, averaged over the five benchmarks. Across all three models, \ouralg{} better preserves BF16 accuracy while keeping generation length close to the original model. For example, on Qwen3-Next-80B-A3B-Thinking, \ouralg{} achieves $81.2\%$ average accuracy versus $76.5$--$77.8\%$ for competing methods, with an average generation length of $15.4$K tokens compared with $15.0$K for BF16. These results indicate that \ouralg{} mitigates both accuracy degradation and quantization-induced overthinking.

\subsection{Inference Efficiency}
\label{sec:efficiency}

\begin{table}[t]
\caption{Inference efficiency of \ouralg{} (2.02 bits) on Qwen3.6-35B-A3B, measured on a single NVIDIA RTX PRO 6000 GPU.}
\label{tab:efficiency}
\vspace{-0.2in}
\begin{center}
\small
\setlength{\tabcolsep}{6pt}
% \begin{tabular}{lcc}
\begin{tabularx}{\textwidth}{l*{3}{>{\centering\arraybackslash}X}}
\toprule
Metric & \ouralg{} & Relative to BF16 \\
\midrule
Routed-expert memory                  & 10.3\,GB          & 6.3$\times$ smaller (64.4\,GB) \\
KV-cache capacity                     & --                & 3.8$\times$ larger \\
MoE-layer latency (batch $\geq 16$)   & --                & 4.0--4.3$\times$ faster \\
Decoding throughput (batch 48)        & 3{,}182 tok/s     & 2.09$\times$ \\
Decoding throughput (batch 128)       & 4{,}880 tok/s     & 1.92$\times$ \\
\bottomrule
\end{tabularx}
\end{center}
\end{table}

Low-bit routed experts reduce both memory footprint and decoding latency.
\Cref{tab:efficiency} reports the inference efficiency of \ouralg{} on Qwen3.6-35B-A3B with a single RTX PRO 6000 GPU.
\ouralg{} shrinks routed-expert memory from 64.4\,GB to 10.3\,GB, leaving 3.8$\times$ more memory for the KV cache.
For batch sizes of 16 or more, the mixed-precision MoE layer runs 4.0--4.3$\times$ faster than BF16, and end-to-end decoding throughput improves by 2.09$\times$ at batch 48 and 1.92$\times$ at batch 128.
The end-to-end gain is smaller than the MoE-layer gain because the non-expert components remain in BF16.
Since \ouralg{} also keeps generation length close to BF16 (\cref{fig:pareto}), these per-token gains are not offset by longer reasoning traces.
Per-level MoE-layer latencies are reported in \cref{app:efficiency}.

\section{Ablation Studies}
\label{sec:ablation}

\subsection{Bit Allocation Across Models}
\label{sec:abl-alloc}

\begin{figure}[t]
  \centering
  % \newlength{\bitallocH}\setlength{\bitallocH}{0.334\linewidth}% panel height
  % \newlength{\bitallocGap}\setlength{\bitallocGap}{0.04\linewidth}% gap between panels
  % \includegraphics[height=\bitallocH,trim=0 0 726.8 0,clip]{figures/bit_allocation.pdf}\hspace{\bitallocGap}%
  % \includegraphics[height=\bitallocH,trim=418 0 363.8 0,clip]{figures/bit_allocation.pdf}\hspace{\bitallocGap}%
  % \includegraphics[height=\bitallocH,trim=781 0 0 0,clip]{figures/bit_allocation.pdf}
  \includegraphics[width=\linewidth]{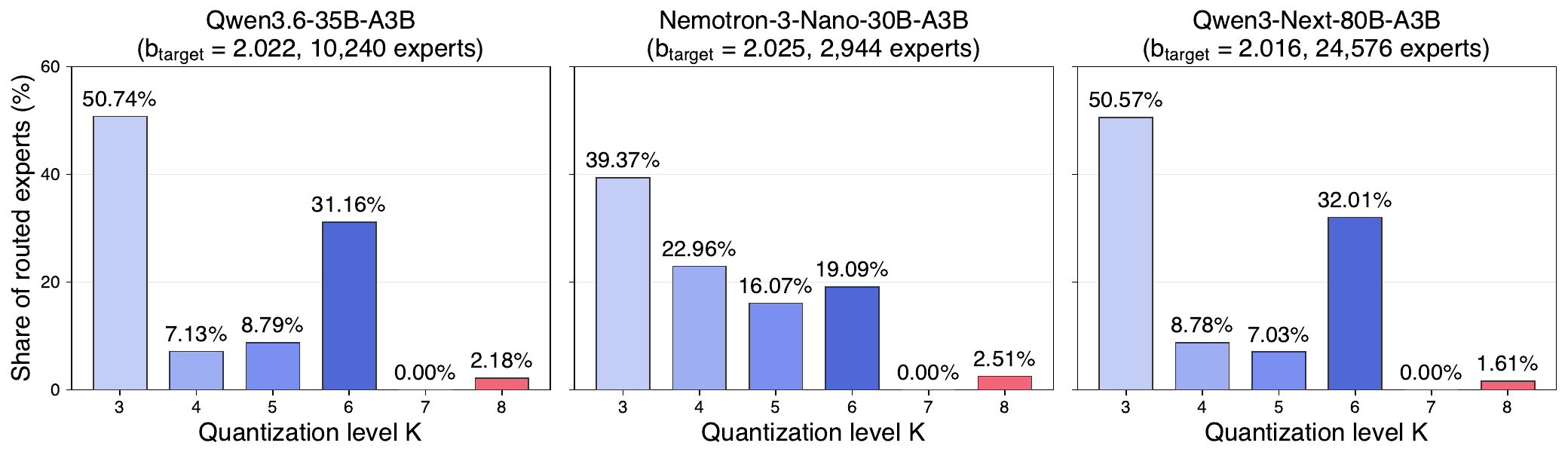}
  \vspace{-0.2in}
\caption{Percentage of routed experts assigned to each quantization level $K \in \{3,4,5,6,8\}$ by RTAQ. The red bars indicate the termination-aware experts, which are fixed at $K{=}8$.}
  \label{fig:bit-allocation}
\end{figure}

Figure~\ref{fig:bit-allocation} shows the resulting bit allocation. Although the knapsack algorithm assigns each expert its own level, it consistently places the largest share at the lowest one, i.e., $K{=}3$ (1.58 bits): 50.7\%, 39.4\% and 50.6\% of the routed experts of Qwen3.6-35B-A3B, Nemotron-3-Nano-30B-A3B and Qwen3-Next-80B-A3B, respectively. The next most common level on both Qwen models is $K{=}6$ (31.2\% and 32.0\%). Levels above this rarely provide meaningful benefits. On Qwen3.6-35B-A3B, allowing up to $K{=}9$ lowers the predicted cost but raises the held-out NLL (0.4707 vs.\ 0.4672).
The termination-aware experts are the exception. They make up only 2.18\%, 2.51\% and 1.61\% of the routed experts and are held at $K{=}8$, which adds 0.016--0.025 bits to the 2-bit average of the remaining experts. \Cref{sec:abl-causal} shows that conventional sensitivity criteria would not protect them, and \cref{sec:abl-selection} shows that protecting them, rather than other experts, is what improves termination.

\subsection{Termination-Aware Experts and Conventional Sensitivity Criteria}
\label{sec:abl-causal}

\begin{figure}[t]
  \centering
  \begin{subfigure}[b]{0.455\linewidth}
    \centering
    \includegraphics[width=\linewidth]{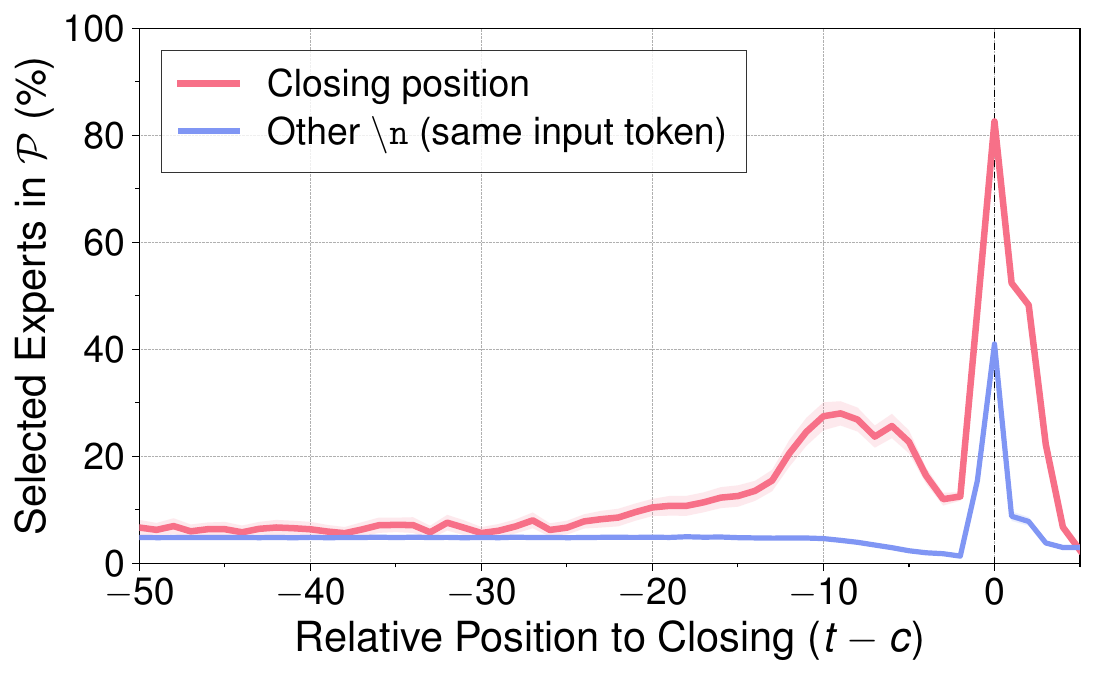}
    \vspace{-5mm}
    \caption{}
    \label{fig:te-event-aligned}
  \end{subfigure}\hfill
  \begin{subfigure}[b]{0.525\linewidth}
    \centering
    \includegraphics[width=\linewidth]{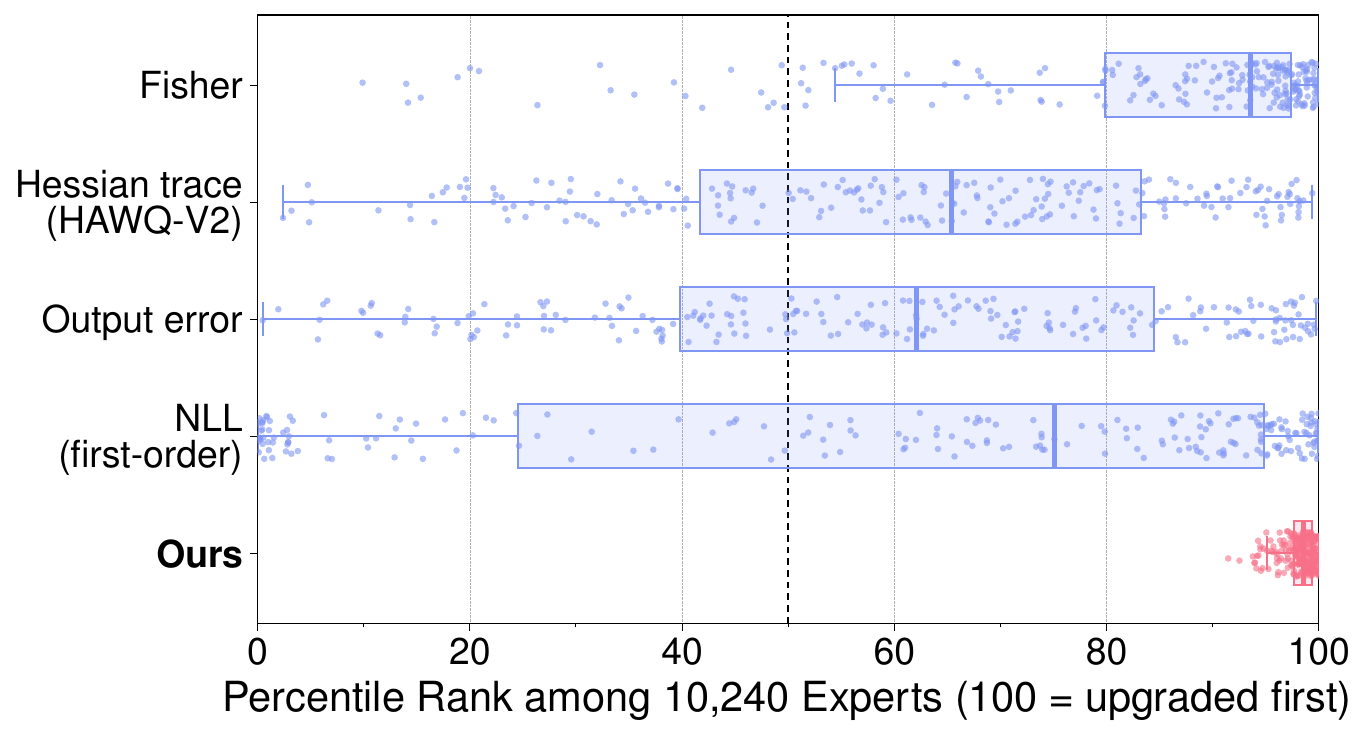}
    \vspace{-5mm}
    \caption{}
    \label{fig:te-sensitivity}
  \end{subfigure}
  \vspace{-2mm}
\caption{\texttt{</think>}-token-specific experts $\mathcal{P}$ in Qwen3.6-35B-A3B. (a) Percentage of the experts in $\mathcal{P}$ selected by the router at each token. Red: tokens just before the model ends its reasoning with \texttt{</think>} (0 on the x-axis). Grey: tokens around ordinary line breaks in the middle of the reasoning, shown for comparison. Most of $\mathcal{P}$ is selected only at the end of reasoning (373 responses from AIME 2025+2026 and GPQA-Diamond; shading: 95\% CI). (b) How important each sensitivity criterion considers the experts in $\mathcal{P}$, as a rank among all 10,240 routed experts (100 = most important). Conventional criteria rank many of them in the middle or lower half, whereas Ours ranks all of them near the top. Dashed line: a random expert.}

  \label{fig:te-analysis}
\end{figure}

To verify that the \texttt{</think>}-token-specific experts $\mathcal{P}$ form a distinct set, we measure on Qwen3.6-35B-A3B the share of $\mathcal{P}$ selected at each position of a response, aligned to its closing position. As shown in Figure~\ref{fig:te-event-aligned}, about 83\% of $\mathcal{P}$ is selected at the closing position, against 5--7\% during reasoning and 41\% at other newlines, the input token at every closing position. Their activation thus follows the decision to stop rather than the input token.

We then examine whether sensitivity criteria used in mixed-precision quantization would protect these experts. As shown in Figure~\ref{fig:te-sensitivity}, the Hessian trace, output error and NLL rank $\mathcal{P}$ near the middle (medians 62--75), and even the Fisher, which ranks them higher (median 94), leaves 39\% outside its top 10\%. Averaged over ordinary tokens, these criteria see $\mathcal{P}$ only at the few positions where it is decisive, and so would leave many of these experts at low bit-widths. RTAQ instead identifies $\mathcal{P}$ directly from routing statistics and holds it at $K{=}8$.

\subsection*{AI use statement}

We used generative AI tools to assist with figure and table editing, and routine coding. AI-assisted code was reviewed and tested by the authors, and all reported results were independently produced and verified by the authors. All AI-assisted text and figures were reviewed against the underlying experiments and data. The authors take full responsibility for the content of this paper.

\bibliography{references}

@article{frantar2022gptq,
  title={Gptq: Accurate post-training quantization for generative pre-trained transformers},
  author={Frantar, Elias and Ashkboos, Saleh and Hoefler, Torsten and Alistarh, Dan},
  journal={arXiv preprint arXiv:2210.17323},
  year={2022}
}

@article{dadgarnia2026gsq,
  title={GSQ: Highly-Accurate Low-Precision Scalar Quantization for LLMs via Gumbel-Softmax Sampling},
  author={Dadgarnia, Alireza and Tabesh, Soroush and Nikdan, Mahdi and Helcig, Michael and Kurtic, Eldar and Kleinegger, Maximilian and Alistarh, Dan},
  journal={arXiv preprint arXiv:2604.18556},
  year={2026}
}

@article{wang2026attend,
  title={Attend to Your Own Thoughts: Breaking the Barrier for Post-Training Quantization of Reasoning LLMs through the Lens of 1.58-Bit Quantization},
  author={Wang, Shigeng and Li, Chao and Kang, Yangyuxuan and Fan, Jiawei and Yao, Anbang},
  journal={arXiv preprint arXiv:2608.01078},
  year={2026}
}

@article{lian2026quantization,
  title={Quantization Inflates Reasoning: Token Inflation as a Hidden Cost of Low-Bit Reasoning Models},
  author={Lian, Xinyu and Krichene, Walid and Huang, Beichen and Tanaka, Masahiro and Ruwase, Olatunji and Zhang, Li and Zhang, Minjia},
  journal={arXiv preprint arXiv:2606.25519},
  year={2026}
}

@inproceedings{
    okoshi2026towards,
    title={Towards Quantization-Aware Training for Ultra-Low-Bit Reasoning {LLM}s},
    author={Yasuyuki Okoshi and Hikari Otsuka and Daichi Fujiki and Masato Motomura},
    booktitle={The Fourteenth International Conference on Learning Representations},
    year={2026},
    url={https://openreview.net/forum?id=Azsd2qyK6C}
}

@article{lotfi2026quantized,
  title={Quantized reasoning models think they need to think longer, but they do not},
  author={Lotfi, Sanae and Kirichenko, Polina and Li, Steven and Liu, Zechun},
  journal={arXiv preprint arXiv:2606.00206},
  year={2026}
}

@inproceedings{liu2025quantization,
    title={Quantization Hurts Reasoning? An Empirical Study on Quantized Reasoning Models},
    author={Ruikang Liu and Yuxuan Sun and Manyi Zhang and Haoli Bai and Xianzhi Yu and Tiezheng YU and Chun Yuan and Lu Hou},
    booktitle={Second Conference on Language Modeling},
    year={2025},
    url={https://openreview.net/forum?id=BM192Ps5Nv}
}

@inproceedings{
    zhang2026when,
    title={When Reasoning Meets Compression: Understanding the Effects of {LLM}s Compression on Large Reasoning Models},
    author={Nan Zhang and Eugene Kwek and Yusen Zhang and Ngoc-Hieu Nguyen and Prasenjit Mitra and Rui Zhang},
    booktitle={The Fourteenth International Conference on Learning Representations},
    year={2026},
    url={https://openreview.net/forum?id=2za3iNkwXn}
}

@inproceedings{
    huang2025mixture,
    title={Mixture Compressor for Mixture-of-Experts {LLM}s Gains More},
    author={Wei Huang and Yue Liao and Jianhui Liu and Ruifei He and Haoru Tan and Shiming Zhang and Hongsheng Li and Si Liu and XIAOJUAN QI},
    booktitle={The Thirteenth International Conference on Learning Representations},
    year={2025},
    url={https://openreview.net/forum?id=hheFYjOsWO}
}

@article{chowdhury2026efficient,
  title={Efficient Quantization of Mixture-of-Experts with Theoretical Generalization Guarantees},
  author={Chowdhury, Mohammed Nowaz Rabbani and El Maghraoui, Kaoutar and Tsai, Hsinyu and Wang, Naigang and Burr, Geoffrey W. and Liu, Liu and Wang, Meng},
  journal={arXiv preprint arXiv:2604.06515},
  year={2026}
}

@inproceedings{duanmu2025mxmoe,
  title={MxMoE: Mixed-precision Quantization for MoE with Accuracy and Performance Co-Design},
  author={Duanmu, Haojie and Li, Xiuhong and Yuan, Zhihang and Zheng, Size and Duan, Jiangfei and Zhang, Xingcheng and Lin, Dahua},
  booktitle={International Conference on Machine Learning},
  year={2025}
}

@inproceedings{chittyvenkata2025mopeq,
  title={MoPEQ: Mixture of Mixed Precision Quantized Experts},
  author={Chitty-Venkata, Krishna Teja and Ye, Jie and Emani, Murali},
  booktitle={Proceedings of the IEEE/CVF International Conference
             on Computer Vision Workshops},
  pages={4082--4091},
  year={2025}
}

@article{alimaskina2026extreme,
  title={Extreme Low-Bit Inference in Reasoning Models: Failure Modes and Targeted Recovery},
  author={Alimaskina, Ekaterina and Rudas, Darya and Shveykin, Denis and Molodtsov, Gleb and Vasiliev, Pavel and Beznosikov, Aleksandr},
  journal={arXiv preprint arXiv:2606.02011},
  year={2026}
}

@article{liu2024deepseek,
  title={Deepseek-v3 technical report},
  author={Liu, Aixin and Feng, Bei and Xue, Bing and Wang, Bingxuan and Wu, Bochao and Lu, Chengda and Zhao, Chenggang and Deng, Chengqi and Zhang, Chenyu and Ruan, Chong and others},
  journal={arXiv preprint arXiv:2412.19437},
  year={2024}
}

@article{zeng2026glm,
  title={Glm-5: from vibe coding to agentic engineering},
  author={Zeng, Aohan and Lv, Xin and Hou, Zhenyu and Du, Zhengxiao and Zheng, Qinkai and Chen, Bin and Yin, Da and Ge, Chendi and Huang, Chenghua and Xie, Chengxing and others},
  journal={arXiv preprint arXiv:2602.15763},
  year={2026}
}

@article{fisher1958grouping,
  title   = {On grouping for maximum homogeneity},
  author  = {Fisher, Walter D.},
  journal = {Journal of the American Statistical Association},
  volume  = {53}, number = {284}, pages = {789--798}, year = {1958}
}

@article{bellman1973note,
  title   = {A note on cluster analysis and dynamic programming},
  author  = {Bellman, Richard},
  journal = {Management Science},
  volume  = {19}, number = {11}, pages = {1328--1329}, year = {1973}
}

@article{aloise2009np,
  title   = {{NP}-hardness of {E}uclidean sum-of-squares clustering},
  author  = {Aloise, Daniel and Deshpande, Amit and Hansen, Pierre and Popat, Preyas},
  journal = {Machine Learning},
  volume  = {75}, number = {2}, pages = {245--248}, year = {2009}
}

@article{mahajan2012planar,
  title   = {The planar $k$-means problem is {NP}-hard},
  author  = {Mahajan, Meena and Nimbhorkar, Prajakta and Varadarajan, Kasturi},
  journal = {Theoretical Computer Science},
  volume  = {442}, pages = {13--21}, year = {2012}
}

@inproceedings{shazeer2017outrageously,
  title     = {Outrageously Large Neural Networks: The Sparsely-Gated Mixture-of-Experts Layer},
  author    = {Shazeer, Noam and Mirhoseini, Azalia and Maziarz, Krzysztof and Davis, Andy and Le, Quoc and Hinton, Geoffrey and Dean, Jeff},
  booktitle = {International Conference on Learning Representations (ICLR)},
  year      = {2017}
}

@article{fedus2022switch,
  title   = {Switch Transformers: Scaling to Trillion Parameter Models with Simple and Efficient Sparsity},
  author  = {Fedus, William and Zoph, Barret and Shazeer, Noam},
  journal = {Journal of Machine Learning Research},
  volume  = {23},
  number  = {120},
  pages   = {1--39},
  year    = {2022}
}

@article{yang2025,
  title   = {{Qwen3} Technical Report},
  author  = {Yang, An and others},
  journal = {arXiv preprint arXiv:2505.09388},
  year    = {2025}
}

@inproceedings{han2016deep,
  title     = {Deep Compression: Compressing Deep Neural Networks with Pruning, Trained Quantization and Huffman Coding},
  author    = {Han, Song and Mao, Huizi and Dally, William J.},
  booktitle = {International Conference on Learning Representations (ICLR)},
  year      = {2016}
}

@inproceedings{kim2024,
  title     = {{SqueezeLLM}: Dense-and-Sparse Quantization},
  author    = {Kim, Sehoon and Hooper, Coleman and Gholami, Amir and Dong, Zhen and Li, Xiuyu and Shen, Sheng and Mahoney, Michael W. and Keutzer, Kurt},
  booktitle = {International Conference on Machine Learning (ICML)},
  year      = {2024}
}

@article{li2016ternary,
  title   = {Ternary Weight Networks},
  author  = {Li, Fengfu and Zhang, Bo and Liu, Bin},
  journal = {arXiv preprint arXiv:1605.04711},
  year    = {2016}
}

@article{ma2024bitnet,
  title   = {The Era of 1-bit {LLMs}: All Large Language Models are in 1.58 Bits},
  author  = {Ma, Shuming and Wang, Hongyu and Ma, Lingxiao and Wang, Lei and Wang, Wenhui and Huang, Shaohan and Dong, Li and Wang, Ruiping and Xue, Jilong and Wei, Furu},
  journal = {arXiv preprint arXiv:2402.17764},
  year    = {2024}
}

@inproceedings{nagel2020adaround,
  title     = {Up or Down? {Adaptive} Rounding for Post-Training Quantization},
  author    = {Nagel, Markus and Amjad, Rana Ali and van Baalen, Mart and Louizos, Christos and Blankevoort, Tijmen},
  booktitle = {International Conference on Machine Learning (ICML)},
  year      = {2020}
}

@inproceedings{frantar2023gptq,
  title     = {{GPTQ}: Accurate Post-Training Quantization for Generative Pre-trained Transformers},
  author    = {Frantar, Elias and Ashkboos, Saleh and Hoefler, Torsten and Alistarh, Dan},
  booktitle = {International Conference on Learning Representations (ICLR)},
  year      = {2023}
}

@inproceedings{dong2019hawq,
  title     = {{HAWQ}: Hessian {AWare} Quantization of Neural Networks with Mixed-Precision},
  author    = {Dong, Zhen and Yao, Zhewei and Gholami, Amir and Mahoney, Michael W. and Keutzer, Kurt},
  booktitle = {IEEE/CVF International Conference on Computer Vision (ICCV)},
  year      = {2019}
}

@inproceedings{dong2020hawqv2,
  title     = {{HAWQ-V2}: Hessian Aware trace-Weighted Quantization of Neural Networks},
  author    = {Dong, Zhen and Yao, Zhewei and Arfeen, Daiyaan and Gholami, Amir and Mahoney, Michael W. and Keutzer, Kurt},
  booktitle = {Advances in Neural Information Processing Systems (NeurIPS)},
  year      = {2020}
}

@inproceedings{wang2019haq,
  title     = {{HAQ}: Hardware-Aware Automated Quantization with Mixed Precision},
  author    = {Wang, Kuan and Liu, Zhijian and Lin, Yujun and Lin, Ji and Han, Song},
  booktitle = {IEEE/CVF Conference on Computer Vision and Pattern Recognition (CVPR)},
  year      = {2019}
}

@article{martens2020new,
  title   = {New Insights and Perspectives on the Natural Gradient Method},
  author  = {Martens, James},
  journal = {Journal of Machine Learning Research},
  volume  = {21},
  number  = {146},
  pages   = {1--76},
  year    = {2020}
}

@inproceedings{kunstner2019limitations,
  title     = {Limitations of the Empirical {Fisher} Approximation for Natural Gradient Descent},
  author    = {Kunstner, Frederik and Balles, Lukas and Hennig, Philipp},
  booktitle = {Advances in Neural Information Processing Systems (NeurIPS)},
  year      = {2019}
}

@article{sinha1979multiple,
  title   = {The Multiple-Choice Knapsack Problem},
  author  = {Sinha, Prabhakant and Zoltners, Andris A.},
  journal = {Operations Research},
  volume  = {27},
  number  = {3},
  pages   = {503--515},
  year    = {1979}
}

@article{everett1963generalized,
  title   = {Generalized {Lagrange} Multiplier Method for Solving Problems of Optimum Allocation of Resources},
  author  = {Everett, Hugh},
  journal = {Operations Research},
  volume  = {11},
  number  = {3},
  pages   = {399--417},
  year    = {1963}
}

@article{shoham1988efficient,
  title   = {Efficient Bit Allocation for an Arbitrary Set of Quantizers},
  author  = {Shoham, Yair and Gersho, Allen},
  journal = {IEEE Transactions on Acoustics, Speech, and Signal Processing},
  volume  = {36},
  number  = {9},
  pages   = {1445--1453},
  year    = {1988}
}

@article{lloyd1982least,
  title   = {Least Squares Quantization in {PCM}},
  author  = {Lloyd, Stuart P.},
  journal = {IEEE Transactions on Information Theory},
  volume  = {28},
  number  = {2},
  pages   = {129--137},
  year    = {1982}
}

@article{wang2011ckmeans,
  title   = {{Ckmeans.1d.dp}: Optimal $k$-means Clustering in One Dimension by Dynamic Programming},
  author  = {Wang, Haizhou and Song, Mingzhou},
  journal = {The R Journal},
  volume  = {3},
  number  = {2},
  pages   = {29--33},
  year    = {2011}
}

@article{gronlund2017fast,
  title   = {Fast Exact $k$-Means, $k$-Medians and {Bregman} Divergence Clustering in 1{D}},
  author  = {Gr{\o}nlund, Allan and Larsen, Kasper Green and Mathiasen, Alexander and Nielsen, Jesper Sindahl and Schneider, Stefan and Song, Mingzhou},
  journal = {arXiv preprint arXiv:1701.07204},
  year    = {2017}
}

@article{openai2024,
  title   = {{OpenAI} o1 System Card},
  author  = {{OpenAI}},
  journal = {arXiv preprint arXiv:2412.16720},
  year    = {2024}
}

@article{guo2025deepseek,
  title   = {{DeepSeek-R1}: Incentivizing Reasoning Capability in {LLMs} via Reinforcement Learning},
  author  = {Guo, Daya and others},
  journal = {arXiv preprint arXiv:2501.12948},
  year    = {2025}
}

@article{chen2024overthinking,
  title   = {Do {NOT} Think That Much for 2+3=? {On} the Overthinking of o1-Like {LLMs}},
  author  = {Chen, Xingyu and Xu, Jiahao and Liang, Tian and He, Zhiwei and Pang, Jianhui and Yu, Dian and Song, Linfeng and Liu, Qiuzhi and Zhou, Mengfei and Zhang, Zhuosheng and Wang, Rui and Tu, Zhaopeng and Mi, Haitao and Yu, Dong},
  journal = {arXiv preprint arXiv:2412.21187},
  year    = {2024}
}

@article{sui2025,
  title   = {Stop Overthinking: A Survey on Efficient Reasoning for Large Language Models},
  author  = {Sui, Yang and Chuang, Yu-Neng and Wang, Guanchu and Zhang, Jiamu and Zhang, Tianyi and Yuan, Jiayi and Liu, Hongyi and Wen, Andrew and Zhong, Shaochen and Chen, Hanjie and Hu, Xia},
  journal = {arXiv preprint arXiv:2503.16419},
  year    = {2025}
}

@article{chen2026minimax,
  title={The minimax-m2 series: Mini activations unleashing max real-world intelligence},
  author={Chen, Aili and Li, Aonian and Zhou, Baichuan and Gong, Bangwei and Jiang, Binyang and Dan, Boji and Zhang, Changhao and Yu, Changqing and Wang, Chao and Ma, Cheng and others},
  journal={arXiv preprint arXiv:2605.26494},
  year={2026}
}

@article{yang2025qwen3,
  title={Qwen3 technical report},
  author={Yang, An and Li, Anfeng and Yang, Baosong and Zhang, Beichen and Hui, Binyuan and Zheng, Bo and Yu, Bowen and Gao, Chang and Huang, Chengen and Lv, Chenxu and others},
  journal={arXiv preprint arXiv:2505.09388},
  year={2025}
}

@article{frantar2023qmoe,
  title={Qmoe: Practical sub-1-bit compression of trillion-parameter models},
  author={Frantar, Elias and Alistarh, Dan},
  journal={arXiv preprint arXiv:2310.16795},
  year={2023}
}

@article{kim2023mixture,
  title={Mixture of quantized experts (moqe): Complementary effect of low-bit quantization and robustness},
  author={Kim, Young Jin and Fahim, Raffy and Awadalla, Hany Hassan},
  journal={arXiv preprint arXiv:2310.02410},
  year={2023}
}

@InProceedings{Chitty-Venkata_2025_ICCV,
    author    = {Chitty-Venkata, Krishna Teja and Ye, Jie and Emani, Murali},
    title     = {MoPEQ: Mixture of Mixed Precision Quantized Experts},
    booktitle = {Proceedings of the IEEE/CVF International Conference on Computer Vision (ICCV) Workshops},
    month     = {October},
    year      = {2025},
    pages     = {4082-4091}
}

@misc{
    li2024examining,
    title={Examining Post-Training Quantization for Mixture-of-Experts: A Benchmark},
    author={Pingzhi Li and Xiaolong Jin and Yu Cheng and Tianlong Chen},
    year={2024},
    url={https://openreview.net/forum?id=sMwYn2lZjO}
}

@inproceedings{gu2026tileq,
  title={TileQ: Efficient Low-Rank Quantization of Mixture-of-Experts with 2D Tiling},
  author={Gu, Hongyaoxing and Chen, Xinzhe and Hu, Lijuan and Liu, Fangfang},
  booktitle={Forty-third International Conference on Machine Learning},
  year={2026},
  url={https://openreview.net/forum?id=ReSBUj2514}
}

@inproceedings{
    deng2026gemq,
    title={{GEMQ}: Global Expert-Level Mixed-Precision Quantization for MoE {LLM}s},
    author={Jianing Deng and Song Wang and Dongwei Wang and Zijie Liu and Tianlong Chen and Huanrui Yang and Jingtong Hu},
    booktitle={Forty-third International Conference on Machine Learning},
    year={2026},
    url={https://openreview.net/forum?id=5Q4ts0pr9T}
}

@article{hooper2025fgmp,
  title={FGMP: fine-grained mixed-precision weight and activation quantization for hardware-accelerated LLM inference},
  author={Hooper, Coleman and Sakr, Charbel and Keller, Ben and Venkatesan, Rangharajan and Keutzer, Kurt and Shao, Sophia and Khailany, Brucek},
  journal={arXiv preprint arXiv:2504.14152},
  year={2025}
}

@misc{autoquantize,
  title={AutoQuantize: A Fast Automatic Mixed-Precision Assignment},
  author={NVIDIA, Model Optimizer Team},
  url={https://nvidia.github.io/Model-Optimizer/announcements/autoquantize.html},
}

@inproceedings{shen2020q,
  title={Q-bert: Hessian based ultra low precision quantization of bert},
  author={Shen, Sheng and Dong, Zhen and Ye, Jiayu and Ma, Linjian and Yao, Zhewei and Gholami, Amir and Mahoney, Michael W and Keutzer, Kurt},
  booktitle={Proceedings of the AAAI conference on artificial intelligence},
  volume={34},
  number={05},
  pages={8815--8821},
  year={2020}
}

@article{cobbe2021training,
  title={Training verifiers to solve math word problems},
  author={Cobbe, Karl and Kosaraju, Vineet and Bavarian, Mohammad and Chen, Mark and Jun, Heewoo and Kaiser, Lukasz and Plappert, Matthias and Tworek, Jerry and Hilton, Jacob and Nakano, Reiichiro and others},
  journal={arXiv preprint arXiv:2110.14168},
  year={2021}
}

@misc{aime25,
  title={American Invitational Mathematics Examination (AIME) 2025}, 
  author={Zhang, Yifan and Math-AI, Team},
  year={2025},
}

@article{rein2023gpqa,
  title={Gpqa: A graduate-level google-proof q\&a benchmark},
  author={Rein, David and Hou, Betty Li and Stickland, Asa Cooper and Petty, Jackson and Pang, Richard Yuanzhe and Dirani, Julien and Michael, Julian and Bowman, Samuel R},
  journal={arXiv preprint arXiv:2311.12022},
  year={2023}
}

@inproceedings{jain2025livecodebench,
  title={Livecodebench: Holistic and contamination free evaluation of large language models for code},
  author={Jain, Naman and Gu, Alex and Li, Wen-Ding and Yan, Fanjia and Zhang, Tianjun and Wang, Sida and Solar-Lezama, Armando and Sen, Koushik and Stoica, Ion},
  booktitle={International Conference on Learning Representations},
  volume={2025},
  pages={58791--58831},
  year={2025}
}
\bibliographystyle{iclr2027_conference}

\newpage
\appendix
\section{Appendix}
\subsection{Calibration Sets}
\label{app:calibration}

Every stage of the method that touches data uses the same calibration corpus: the second moments of the codebook
bank (Stage~1), the Fisher costs of the allocation (Stage~2), and the Hessians and held-out damping selection of the
GPTQ refinement (Stage~3). The baselines we compare against are given the identical corpus, so a difference between
methods is never a difference in calibration data.

\paragraph{Stratified candidate pool.}
We first sample a candidate pool eight times larger than the final set, stratified by source so that the pool cannot
drift toward whichever domain happens to be most numerous upstream. The pool holds $1{,}024$ documents:

\begin{center}
\vspace{-0.1in}
\small
% \begin{tabular}{lrr}
\begin{tabularx}{\textwidth}{l*{3}{>{\centering\arraybackslash}X}}
\toprule
Source & Pool & Selected \\
\midrule
\texttt{openr1\_math\_verified}       & 256 & 32 \\
\texttt{nemotron\_ptd\_math}          & 128 & 16 \\
\texttt{nemotron\_math\_v3\_verified} & 128 & 16 \\
\texttt{nemotron\_ptd\_code}          & 128 & 16 \\
\texttt{nemotron\_ptd\_stem}          & 128 & 16 \\
\texttt{nemotron\_ptd\_chat}          &  64 &  8 \\
\texttt{nemotron\_science}            &  64 &  8 \\
\texttt{nemotron\_chat\_reasoning\_off} & 64 &  8 \\
\texttt{cnn\_dailymail\_anchor}       &  64 &  8 \\
\midrule
Total & 1{,}024 & 128 \\
\bottomrule
\end{tabularx}
\end{center}

\paragraph{Coverage-greedy selection.}
Which $128$ of the $1{,}024$ documents are kept is decided by routed-expert coverage rather than at random. We run a
BF16 forward pass over the whole pool, recording for every document the number of times each (layer, expert) unit is
selected by the router, and then fill the per-source quotas greedily under a concave utility. With $c_d \in
\mathbb{R}^{L \times E}$ the routing counts of document $d$ and $t$ the counts accumulated by the documents already
chosen, the next document from a source with quota remaining is
\[
  d^\star \;=\; \arg\max_{d}\; \big\langle c_d,\; (t + 1)^{-1/2} \big\rangle .
\]
The $(t+1)^{-1/2}$ weighting makes an expert worth less once it is already well covered, so the selection spends its
budget on rarely routed experts instead of re-sampling the popular ones. This matters because the cost
sums over the tokens routed to an expert: an expert that no calibration token reaches has no cost estimate at all,
and the knapsack would assign it a level on no evidence.

\paragraph{Packing and the loss mask.}
Selected documents are concatenated into rows of exactly $2{,}048$ tokens; the tail that does not fill a row is
dropped. The Fisher gradients of Stage~2 are taken with respect to the mean loss over \emph{assistant} tokens only
(the reasoning and answer tokens of the assistant turn plus its closing turn token), computed under a full-context
forward pass. Tokens preceding the first assistant turn have no determinable role and are excluded. The Fisher cost
is estimated from the first $16$ rows, which additionally carry the per-token gradient at each MoE block output,
except where noted below.

\paragraph{On-policy construction.}
The assistant turns of the source datasets were written by other models. Calibrating on that text measures the
quantizer against states the target model does not itself visit, and it describes the target's router poorly.
The prompts of the same $1{,}024$ documents are re-rendered with the target's chat template and re-answered by the
target model in BF16 under its own evaluation sampling protocol; thinking is enabled exactly where the source
document had a thinking block, so a reasoning-off document stays reasoning-off. Document $=$ prompt $+$ response,
truncated at $2{,}048$ tokens, with the loss mask rebuilt over the new assistant span. Only then does the routing
scan run, so both the text and the router that scores coverage belong to the model being quantized.

\paragraph{Held-out split.}
A disjoint set of $128$ documents, sampled from the same sources under the same quotas and capped at $4{,}096$
tokens, is held out. It is used for the candidate comparison that picks the allocation (Stage~2), for the damping
grid of Stage~3, and for convergence and rollback decisions; it is never used for computing the costs themselves,
and never for evaluation. Benchmark numbers come from frozen checkpoints only.

\paragraph{Sets used per model.}

\begin{center}
\small
% \begin{tabular}{lcrrr}
\begin{tabularx}{\textwidth}{l*{5}{>{\centering\arraybackslash}X}}
\toprule
Model & On-policy & Selected documents & Packed rows & Fisher rows \\
\midrule
Qwen3.6-35B-A3B          & -- & 128 & 115 & 16 \\
Qwen3-Next-80B-A3B       & \checkmark & 128 & 120 & 16 \\
Nemotron-3-Nano-30B-A3B  & \checkmark & 128 & 120 & 120 \\
\bottomrule
\end{tabularx}
\end{center}

Every set is built by the same procedure from the same pool and the same quotas; only the model that writes the
responses and scores the routing coverage differs. For Nemotron-3-Nano-30B-A3B the Fisher cost is estimated from
all 120 rows: with 16 rows its expert ranking had a split-half rank reliability of only 0.47, which rose to 0.69
with 120 rows.

\section{Supplementary Preliminaries}\label{app:reuse_prelim}

\subsection{Mixed-precision Quantization}
The objective in \cref{eq:layerwise_objective} separates over rows,
\begin{equation}
    \mathcal{E}(\widehat{W})
    =
    \sum_i
    (w_i-\widehat{w}_i)^{\top} S (w_i-\widehat{w}_i),
    \label{eq:rowwise_objective}
\end{equation}
where $w_i$ and $\widehat{w}_i$ denote the $i$-th rows of $W$ and
$\widehat{W}$. It is quadratic in $\widehat{W}$ but combinatorial in the
discrete codes: exact minimization over all $K^{np}$ code assignments is
intractable. Practical methods therefore decompose the problem into codebook
construction, code assignment, and affine-parameter optimization, or optimize
the codes greedily (Section~\ref{sec:method}).

\subsection{Exact One-dimensional $K$-means Clustering} \label{sec:kmeans-appendix}
Learning $K$ quantization levels for a set of scalar values $\{x_b\}$ with
associated masses $\{m_b\}$ can be formulated as the weighted $K$-means problem
\begin{equation}
    \min_{c_1,\ldots,c_K \in \mathbb{R}}
    F(c)
    =
    \sum_b m_b \min_{1 \leq k \leq K} (x_b-c_k)^2 .
    \label{eq:weighted_kmeans}
\end{equation}
In general dimensions, $K$-means is NP-hard
\citep{aloise2009np, mahajan2012planar}, and Lloyd's algorithm
\citep{lloyd1982least}, which alternates nearest-centroid assignment with
centroid updates, converges only to a local optimum that depends on its
initialization. In one dimension, however, the problem admits an exact
polynomial-time solution
\citep{fisher1958grouping, bellman1973note, wang2011ckmeans, gronlund2017fast},
due to the following classical property.

\section{Quantization-Bank Details}\label{app:reuse_bank}
We pool all $np$ normalized values of a matrix into a histogram of
$B=2048$ bins over their $[10^{-4},\,1-10^{-4}]$ quantile range (values
outside are clamped), so that a few outliers cannot stretch the grid.

\paragraph{Gauge fixing.}
\cref{eq:quant_representation} has an affine redundancy:
for any $a\neq0$ and $b$, the substitution
$c_k \mapsto a c_k+b$, $\alpha_i \mapsto \alpha_i/a$,
$\mu_i \mapsto \mu_i-b\alpha_i/a$
leaves every reconstructed weight unchanged.
We remove this redundancy by standardizing the codebook according to
its priors:
\begin{align}
    \bar{c}
    &=
    \sum_{k=1}^{K}\pi_k c_k,
    \qquad
    c_k
    \leftarrow
    \frac{
        c_k-\bar{c}
    }{
        \sqrt{
            \sum_{r=1}^{K}
            \pi_r(c_r-\bar{c})^2
        }
    }.
    \label{eq:gauge_fixing}
\end{align}
This keeps the row scales comparable to the original row standard
deviations and gives every $K$ a consistent storage convention.

\paragraph{The ternary case ($K=3$).}
For three levels, we retain the symmetric alphabet $c=(-1,0,+1)$.
This representation is attractive because symmetric ternary codes can
pack five weights into one byte, admit specialized kernels, and the
per-row affine parameters already capture much of the flexibility of a
learned three-level codebook.

The ternary codes are initialized using the thresholding rule from
ternary weight networks \citep{li2016ternary} and alternated with the
row-wise affine parameters for $15$ iterations:
\begin{align}
    T_{ij}
    &=
    \operatorname{sign}(W_{ij}-\mu_i)
    \,
    \mathbb{I}
    \left[
        |W_{ij}-\mu_i|>\Delta_i
    \right],
    \label{eq:ternary_codes}
    \\
    \Delta_i
    &=
    \frac{0.75}{p}
    \sum_{j=1}^{p}
    |W_{ij}-\mu_i|,
    \label{eq:ternary_threshold}
    \\
    \alpha_i
    &=
    \frac{
        \sum_j T_{ij}(W_{ij}-\mu_i)
    }{
        \sum_j |T_{ij}|
    },
    \label{eq:ternary_scale}
    \\
    \mu_i
    &\leftarrow
    \mu_i
    +
    \frac{1}{p}
    \sum_{j=1}^{p}
    \left(
        W_{ij}-\mu_i-\alpha_iT_{ij}
    \right).
    \label{eq:ternary_offset}
\end{align}

The scale in \cref{eq:ternary_scale} is the least-squares fit
over the non-zero support, while the offset update absorbs the residual
mean.
The ternary variant then proceeds directly to Step 3 with
$\ell_{ij}=T_{ij}$.

\subsection{Step 2: Code Assignment by Coordinate Descent}
\label{sec:stage2_code_assignment}

For $K>3$, after fixing the codebook, we jointly refine the discrete
codes and the row-wise affine parameters by alternating two closed-form
updates for $5$ iterations, in the spirit of Lloyd's algorithm.

For fixed $(\mu_i,\alpha_i)$, each weight is assigned to its nearest
reconstruction level:
\begin{equation}
    q_{ij}
    =
    \arg\min_{1\leq k\leq K}
    \left|
        W_{ij}
        -
        \mu_i
        -
        \alpha_i c_k
    \right|.
    \label{eq:code_assignment}
\end{equation}

Let
\begin{equation}
    \ell_{ij}=c_{q_{ij}},
    \qquad
    \bar{\ell}_i
    =
    \frac{1}{p}
    \sum_j \ell_{ij},
    \qquad
    \bar{W}_i
    =
    \frac{1}{p}
    \sum_j W_{ij}.
\end{equation}
For fixed codes, the least-squares affine parameters are
\begin{align}
    \alpha_i
    &=
    \frac{
        \sum_j
        (\ell_{ij}-\bar{\ell}_i)
        (W_{ij}-\bar{W}_i)
    }{
        \sum_j
        (\ell_{ij}-\bar{\ell}_i)^2
    },
    \label{eq:stage2_scale}
    \\
    \mu_i
    &=
    \bar{W}_i
    -
    \alpha_i\bar{\ell}_i.
    \label{eq:stage2_offset}
\end{align}

Each alternating step exactly minimizes the unweighted row
reconstruction error
\begin{equation}
    \sum_j
    \left(
        W_{ij}
        -
        \mu_i
        -
        \alpha_i\ell_{ij}
    \right)^2
\end{equation}
with respect to one set of variables while holding the other fixed.
The objective is therefore non-increasing, and the iteration converges
to a fixed point.

The assignment in \cref{eq:code_assignment} is deliberately
\emph{unweighted}.
Using the full activation-aware objective would couple all coordinates
of a row through the off-diagonal entries of $S$, making independent
nearest-level assignment invalid.
Instead, activation information enters through the weighted histogram
and the affine relocation.
The fully coupled code assignment is deferred to
Section~\ref{sec:codebook_gptq}, where it is handled with the OBS/GPTQ
machinery after the precision of every expert has been selected.

\subsection{Step 3: Activation-Aware Relocation of the Row Affine Pair}\label{app:reuse_affine}
We freeze the discrete codes and re-optimize
$(\mu_i,\alpha_i)$ against the \emph{full} second-moment matrix $S$.

For row $i$, define
\begin{equation}
    J_i(\mu_i,\alpha_i)
    =
    \left(
        w_i-\mu_i\mathbf{1}-\alpha_i\ell_i
    \right)^{\top}
    S
    \left(
        w_i-\mu_i\mathbf{1}-\alpha_i\ell_i
    \right).
    \label{eq:stage3_objective}
\end{equation}
This is a convex quadratic in the two unknowns
$\alpha_i$ and $\mu_i$.
Setting both partial derivatives to zero gives the normal equations
\begin{equation}
    \begin{bmatrix}
        \ell_i^{\top}S\ell_i
        &
        \ell_i^{\top}S\mathbf{1}
        \\
        \mathbf{1}^{\top}S\ell_i
        &
        \mathbf{1}^{\top}S\mathbf{1}
    \end{bmatrix}
    \begin{bmatrix}
        \alpha_i\\
        \mu_i
    \end{bmatrix}
    =
    \begin{bmatrix}
        \ell_i^{\top}S w_i\\
        \mathbf{1}^{\top}S w_i
    \end{bmatrix}.
    \label{eq:stage3_normal_equations}
\end{equation}

The resulting $2\times2$ systems are solved for all rows at once.
The required quantities can be formed from matrix products involving
$S\ell_i$, $S\mathbf{1}$, and $Sw_i$.

For numerical stability, we regularize the second-moment matrix as
\begin{equation}
    S
    \leftarrow
    S
    +
    \varepsilon I,
    \qquad
    \varepsilon
    =
    \lambda
    \operatorname{mean}
    \left(
        \operatorname{diag}(S)
    \right),
    \qquad
    \lambda=10^{-4}.
    \label{eq:stage3_regularization}
\end{equation}

If the determinant of the $2\times2$ Gram matrix in
\cref{eq:stage3_normal_equations} is smaller than
$\tau=10^{-8}$ times the product of its diagonal entries, the system is
treated as numerically degenerate and the row retains its Step 2
values.

The resulting affine parameters therefore minimize the expert's
activation-weighted output reconstruction error on its routed
calibration inputs while accounting for correlations between input
coordinates.

\section{Sensitivity and Allocation Details}\label{app:reuse_sensitivity}
\subsection{Sensitivity-Aware Mixed-Precision Allocation}
\label{sec:prelim_allocation}

\paragraph{Mixed-precision allocation.}
Not all weights tolerate the same quantization precision. Mixed-precision
methods \citep{dong2019hawq, dong2020hawqv2, wang2019haq} assign a bitwidth to
each block by (i) estimating the increase in task loss $\mathcal{L}$ caused by
quantizing that block and (ii) solving a budget-constrained allocation problem.

\paragraph{Second-order sensitivity.}
Consider a block whose quantization perturbs an intermediate quantity
$u$, either its weights or its output, by $\delta$. A second-order Taylor
expansion of the task loss gives
\begin{equation}
    \Delta \mathcal{L}
    \approx
    g^{\top}\delta
    +
    \frac{1}{2}\delta^{\top}H\delta,
    \qquad
    g
    =
    \frac{\partial \mathcal{L}}{\partial u},
    \qquad
    H
    =
    \frac{\partial^2 \mathcal{L}}{\partial u^2}.
    \label{eq:second_order_sensitivity}
\end{equation}
Around a trained model the first-order term is dropped, on the grounds that the
gradient vanishes at a stationary point of the training loss
\citep{dong2019hawq}, and the quadratic term serves as the sensitivity. Since
the exact Hessian is impractical to compute, mixed-precision methods replace it
with tractable proxies such as the Hessian trace \citep{dong2020hawqv2} or the
\emph{empirical Fisher}
\begin{equation}
    F_{\mathrm{emp}}
    =
    \mathbb{E}\left[gg^{\top}\right],
    \label{eq:empirical_fisher}
\end{equation}
which is constructed from squared gradients and, in practice, further reduced to
its diagonal \citep{martens2020new, kunstner2019limitations}.

Applying \cref{eq:second_order_sensitivity} to
$u=(h_t)_t$ and $\delta=(\delta_t)_t$ relies on the following three
assumptions.

We assume that the BF16 model is sufficiently close to a stationary
point of the training objective such that
$\mathbb{E}[g_t]\approx 0$.
Accordingly, the first-order Taylor term has approximately zero mean
across calibration documents and behaves primarily as finite-sample
noise.
On our calibration set, the summed absolute magnitude of the
first-order term is $2.4\times$ that of the second-order term, while
its signed mean is approximately zero.
Retaining it would therefore tend to rank quantization levels according
to sampling noise.
We omit the first-order term.
Anchoring every expert and every precision level to the same BF16
reference also places all sensitivity costs on a common scale.

\medskip
\noindent
\textbf{Assumption A5 (Diagonal empirical-Fisher curvature).}
\emph{
Let
\begin{equation}
    g_t^{\Sigma}
    =
    \frac{\partial \left(\sum_i \ell_i\right)}
         {\partial h_t}
    =
    T_r g_t
\end{equation}
denote the gradient of the summed loss.
We approximate the curvature of the summed loss at token $t$ by the
empirical Fisher
$g_t^{\Sigma}(g_t^{\Sigma})^{\top}$ and retain only its diagonal.
Because $\mathcal{L}_r$ is an averaged loss, the corresponding
curvature is divided by $T_r$.
}

\medskip
\noindent
\textbf{Assumption A6 (Additivity).}
\emph{
We approximate the loss increase caused by jointly quantizing multiple
experts as the sum of their individual loss increases.
This assumption is what reduces mixed-precision assignment to the
multiple-choice knapsack problem in
\cref{eq:mckp}.
It is valid only within the local second-order approximation, i.e.,
when the individual perturbations remain sufficiently small.
The trust region introduced in
Section~\ref{app:reuse_allocation} is designed to enforce this
condition.
}

\medskip

Substituting Assumption A5 into the second-order Taylor term gives
\begin{equation}
    \frac{1}{2}
    \sum_t
    \delta_t^{\top}
    \frac{
        \operatorname{diag}
        \left(
            g_t^{\Sigma}(g_t^{\Sigma})^{\top}
        \right)
    }{T_r}
    \delta_t
    = \frac{T_i}{2}\sum_{t=1}^{T_i}\sum_{d=1}^{D} g_{t,d}^{2}\,\delta_{t,d}(e,K)^{2}.
\end{equation}
Averaging this quantity over $R_F$ calibration documents yields the
cost
\begin{equation}
    \Delta(e,K)
    =
    \frac{1}{N}
    \sum_{i=1}^{N}
    \frac{T_i}{2}
    \sum_{t \in [T_i]}
    \left\langle
        g_{t}, \delta_{t}
    \right\rangle^2.
    \label{eq:fisher_expert_cost}
\end{equation}

\paragraph{Calibration and Fisher-free alternative.}
We use $N=16$ calibration documents, each containing at most $2048$
tokens, and apply the language-modeling loss only to assistant tokens.
The gradients require one forward--backward pass per calibration
document.
Afterwards, $\Delta(e,K)$ is evaluated for every $K$ by decoding the
corresponding bank entry and executing only the two matrix
multiplications of expert $e$ on its routed tokens.
Thus, constructing the complete cost table is substantially cheaper
than a fine-tuning epoch.

Alongside \cref{eq:fisher_expert_cost}, we also record the
plain routed-output reconstruction error
\begin{equation}
    \Delta_{\mathrm{out}}(e,K)
    =
    \sum_t
    \left\|
        \delta_t
    \right\|_2^2,
    \label{eq:plain_output_cost}
\end{equation}
which provides a Fisher-free alternative for the candidate allocations
described in Section~\ref{app:reuse_allocation}.

\subsection{Knapsack Allocation}\label{app:reuse_allocation}

For a Lagrange multiplier $\lambda\geq0$, the relaxed problem of \cref{eq:mckp} decouples across experts, and each expert independently selects
\begin{equation}
    K_e(\lambda) \in \mathop{\mathrm{argmin}}_{K \in \{3,\dots,6\}} \Delta(e,K) + \lambda K.
    \label{eq:lagrangian_assign}
\end{equation}
The corresponding total quantization level budget is $B(\lambda) = \sum_{e \in \mathcal{A}} K_e(\lambda)$. As $\lambda$ increases, larger codebooks are penalized more heavily, and hence $B(\lambda)$ is non-increasing in $\lambda$. We therefore use bisection to find the smallest $\lambda$ such that $B(\lambda)\leq B$.
The resulting allocation is optimal for the budget it attains. If a small amount of budget remains unused, we greedily spend it by increasing experts in decreasing order of sensitivity reduction per additional quantization level. 
Two classical properties of this relaxation underlie the allocation procedure
of Section~\ref{sec:bit_allocation}.

\medskip
\noindent
\textbf{Lemma 2 (Monotonicity; \citealp{shoham1988efficient}).}
\emph{$\bar{b}(\lambda)$ is non-increasing in $\lambda$.}
\label{lemma:monotonicity}

\medskip
\noindent
\textbf{Proposition 1 (Lagrangian optimality; \citealp{everett1963generalized}).}
\emph{For any $\lambda^\star \geq 0$, the allocation
$\{K_i(\lambda^\star)\}_{i=1}^{N}$ minimizes the total cost of
\cref{eq:mckp} among all allocations whose average bitwidth is at most
$\bar{b}(\lambda^\star)$. In particular, it is feasible for
\cref{eq:mckp} whenever $\bar{b}(\lambda^\star) \leq \bar{b}$.}
\medskip

By Lemma~\ref{lemma:monotonicity}, a suitable $\lambda^\star$ is located by bisection
\citep{shoham1988efficient}, and Proposition~1 certifies optimality for the
discrete budget the resulting allocation actually uses. Because the attainable
average bitwidths are discrete, a small amount of budget may remain unused;
Section~\ref{sec:bit_allocation} describes how this slack is closed.

\paragraph{Trust region.}
The feasible set $\mathcal{K}_{l,e}$ for an expert is not the entire
set $\mathcal{K}$.
\cref{eq:fisher_expert_cost} is a local second-order
approximation whose reliability decreases as the quantization
perturbation grows.
Furthermore, its cost curve as a function of $K$ is not necessarily
convex.
For approximately $6\%$ of routed experts, the cost at the starting
level lies above the chord connecting its two neighboring precision
levels.
An unrestricted knapsack solver could therefore move such experts
toward extreme precisions based on artifacts of the local
approximation.

\paragraph{Outside-budget pinning.}
The pinned precision can be accounted for either \emph{inside} or
\emph{outside} the nominal bit budget.
When pinning is included inside the budget, the remaining
$N-|\mathcal{P}|$ experts are assigned a slightly lower average
precision so that the global target $\bar{b}$ is preserved.

Alternatively, the unpinned experts can retain average precision
$\bar{b}$.
In this case, the resulting average precision over all routed experts
is
\begin{equation}
    \bar{b}'
    =
    \frac{
        \left(N-|\mathcal{P}|\right)\bar{b}
        +
        |\mathcal{P}|\,b_{K_{\mathrm{pin}}}
    }{N}.
    \label{eq:pinned_bitwidth}
\end{equation}
For
$\bar{b}=2$,
$N=10{,}240$,
$|\mathcal{P}|=223$, and
$K_{\mathrm{pin}}=8$,
this gives $\bar{b}' = 2.0218$.

\section{Codebook-Constrained Error Correction Details}\label{app:reuse_correction}
We consider a Transformer where each block consists of layer normalization, attention and sparse Mixture-of-Experts (MoE) architecture with shared experts.

\paragraph{Routing-weighted objective.}
The quantity consumed by the residual stream is the
\emph{routing-weighted} expert output
$w_{t,e}f_e(x_t)$.
For the fused gate/up projection, the squared change in this quantity
over the calibration tokens routed to expert $e$ is
\begin{equation}
    \sum_t
    w_{t,e}^{2}
    \left\|
        \left(
            W_{gu}^{e}
            -
            \widehat{W}_{gu}^{e}
        \right)x_t
    \right\|_2^2
    =
    \operatorname{tr}
    \left[
        \left(
            W_{gu}^{e}
            -
            \widehat{W}_{gu}^{e}
        \right)
        H_{gu}^{e}
        \left(
            W_{gu}^{e}
            -
            \widehat{W}_{gu}^{e}
        \right)^{\top}
    \right],
    \label{eq:routed_gu_objective}
\end{equation}
Thus, the curvature matrix for the row-wise reconstruction objective is
the routing-weight-squared second moment of the expert inputs.
A token with routing weight $0.5$ contributes one quarter as much as a
token with routing weight $1$, reflecting the amount of the expert
output that reaches the residual stream.

\paragraph{Fixed-codebook objective.}
Consider one row $w\in\mathbb{R}^{p}$ with fixed affine parameters
$(\mu,\alpha)$ and an allocated $K$-level codebook.
The code-assignment problem is
\begin{equation}
    \min_{
        q\in\{0,\ldots,K-1\}^{p}
    }
    \left(
        w-\widehat{w}(q)
    \right)^{\top}
    H
    \left(
        w-\widehat{w}(q)
    \right),
    \qquad
    \widehat{w}_{j}(q)
    =
    \mu+\alpha c_{q_j}.
    \label{eq:fixed_codebook_objective}
\end{equation}

Step 2 of the quantization bank solved the corresponding problem with
$H=I$, in which case the objective decouples into independent
nearest-level assignments.
With a full curvature matrix $H$, however, the columns interact:
rounding one coordinate upward can be partly compensated by rounding a
correlated coordinate downward.

\medskip
\noindent
\textbf{Proposition 3 (Format preservation).}
\emph{
The output of the GPTQ sweep decodes using the same codebook
$\{c_k\}_{k=0}^{K-1}$, the same bitwidth, and the same storage layout
as the bank variant selected during allocation.
}

\smallskip
\noindent
\textit{Proof.}
Every stored weight is generated by
\cref{eq:fixed_code_assignment} for some
$q_j\in\{1,\ldots,K\}$ and therefore lies on the reconstruction grid
$\{\mu+\alpha c_k\}_{k=1}^{K}$.
The GPTQ compensation step modifies only columns that have not yet been
quantized and stored.
Hence, neither the code alphabet nor its storage representation is
changed.
\hfill $\square$

\medskip

Three properties are useful in our setting.
First, because \cref{eq:fixed_code_assignment} is applied to
the \emph{updated} column values, a weight may legitimately move to a
neighboring code level in response to errors introduced by previously
quantized columns.
This is precisely the coupled assignment that Step 2 cannot perform;
in our experiments, approximately $13\%$ of the codes change during
this sweep.

Second, when $H=I$, the off-diagonal compensation vanishes, and the
procedure reduces to the independent nearest-level assignment of
\cref{eq:code_assignment}.
Step 2 can therefore be viewed as the special case that ignores
activation correlations.

Third, after the sweep we explicitly verify grid membership
numerically.
Consequently, the resulting codes can be consumed directly by the same
packed inference kernels as the original bank representation.

\subsection[Choosing the damping strength eta]{Choosing the damping strength $\eta$}
\label{sec:damping_selection}

The damping parameter $\eta$ trades fidelity to the calibration
curvature against robustness to unseen inputs.
Small values closely follow the empirical curvature matrix, whereas
large values suppress its anisotropy and move the procedure toward the
uncoupled nearest-level assignment in
\cref{eq:code_assignment}.

We select $\eta$ using calibration data held out from the curvature
fit.
During selection, the final $15$ sequences of the Stage-3 calibration
stream are excluded when constructing $H$.
We evaluate $\eta \in \{0.01,\,0.03,\,0.1,\,0.3,\,1\}$ using the
relative expert-output reconstruction error on these held-out
sequences.

To keep the search inexpensive, damping selection is performed using
the $64$ most frequently routed experts from layers
$0$, $13$, $26$, and $39$.
The selected value of $\eta$ is then used to rebuild all experts using
the full Stage-3 calibration set.

The Stage-3 calibration stream is constructed so that its five source
domains remain approximately token-balanced throughout the stream.
Its held-out tail therefore remains balanced, preventing the damping
selection from disproportionately favoring any single domain.

\paragraph{Affine refit.}
\begin{equation}
    \begin{bmatrix}
        \ell^{\top}H\ell
        &
        \ell^{\top}H\mathbf{1}
        \\
        \mathbf{1}^{\top}H\ell
        &
        \mathbf{1}^{\top}H\mathbf{1}
    \end{bmatrix}
    \begin{bmatrix}
        \alpha\\
        \mu
    \end{bmatrix}
    =
    \begin{bmatrix}
        \ell^{\top}Hw\\
        \mathbf{1}^{\top}Hw
    \end{bmatrix}.
    \label{eq:final_affine_refit}
\end{equation}

The resulting $\mu$ and $\alpha$ are rounded to BF16, matching the
storage format of the quantization bank.
Thus, the deployed weights remain exactly representable by the packed
codebook representation and Proposition~3 continues to hold.
Rows with a numerically degenerate system or a non-positive fitted
scale retain their affine parameters from the bank.

\paragraph{Scalar gain.}
In weight space, the least-squares fits in
Equations~\ref{eq:stage2_scale},
\ref{eq:stage3_normal_equations}, and
\ref{eq:final_affine_refit}
make the reconstruction residual orthogonal to the fitted
representation, leaving little room for an additional scalar
correction at the weight level.
At $K=4$, we observe that the corrected expert outputs have, on
average, approximately $84\%$ of the norm of their BF16 counterparts.
The router nevertheless combines these experts using routing weights
computed under the BF16 model.
The objective is a scalar quadratic,
\begin{equation}
    \gamma^2
    \sum_t\|q_t\|_2^2
    -
    2\gamma
    \sum_t
    \langle q_t,r_t\rangle
    +
    \mathrm{const},
\end{equation}
whose minimizer is exactly
\cref{eq:scalar_gain}.
Experts with fewer than $64$ routed calibration tokens retain
$\gamma_e=1$.
Experts receiving no calibration tokens retain their original bank
variant.
The scalar gain primarily corrects output magnitude rather than
direction; by itself it provides limited improvement, but becomes
effective after the GPTQ reassignment has removed much of the
directional reconstruction error.

\subsection{Computational Cost}
\label{sec:final_stage_cost}

For each expert, this stage constructs two routing-weighted curvature
matrices of size
$2048\times2048$ and $512\times512$, respectively, and factorizes each
matrix once.
It then applies the GPTQ sweep to the $1024$ rows of the fused gate/up
projection and the $2048$ rows of the down projection, solves one
$2\times2$ affine system per row, and computes one scalar output gain.

The resulting correction requires approximately
$0.5$--$0.6$ seconds per expert for the CPU-side optimization.
No GPU is required for the per-expert code-reassignment and affine-fit
steps.
For the full Qwen 3.6 35B A3B model with $10{,}240$ routed experts,
including damping selection and the required calibration passes, the complete stage finishes in about half an hour on 112 CPU cores (calibration passes on GPU). Stage I takes 0.94 hours on four RTX PRO 6000 GPUs and is reused across bit budgets; the Stage II knapsack takes a few seconds on CPU.

\section{Inference Efficiency Details}
\label{app:efficiency}

\Cref{tab:kernel-latency} reports the latency of one MoE layer of Qwen3.6-35B-A3B on a single RTX PRO 6000 GPU when every expert in the layer uses the same codebook size $K$.
Every level is faster than BF16 at all batch sizes, with the largest speedups at 16--128 concurrent tokens.

\begin{table}[h]
\caption{Latency ($\mu$s) of one MoE layer of Qwen3.6-35B-A3B on a single RTX PRO 6000 GPU, where every expert in the layer uses the same codebook size $K$. Parentheses: speedup over BF16.}
\label{tab:kernel-latency}
\vspace{-0.2in}
\begin{center}
\small
\setlength{\tabcolsep}{5pt}
% \begin{tabular}{rcccc}
\begin{tabularx}{\textwidth}{l*{5}{>{\centering\arraybackslash}X}}
\toprule
Concurrent tokens & BF16 & $K=4$ & $K=5$ & $K=6$ \\
\midrule
1   & 41.3   & 22.3 (1.85$\times$)  & 27.1 (1.52$\times$)  & 27.3 (1.51$\times$)  \\
16  & 420.8  & 74.8 (5.63$\times$)  & 106.4 (3.95$\times$) & 104.5 (4.03$\times$) \\
48  & 788.9  & 140.9 (5.60$\times$) & 207.6 (3.80$\times$) & 207.8 (3.80$\times$) \\
128 & 1013.7 & 183.5 (5.52$\times$) & 264.7 (3.83$\times$) & 263.4 (3.85$\times$) \\
256 & 1053.2 & 242.9 (4.34$\times$) & 331.4 (3.18$\times$) & 325.3 (3.24$\times$) \\
\bottomrule
\end{tabularx}
\end{center}
\end{table}

\section{Non-Termination as the Main Failure Mode}
\label{sec:abl-termination}

\begin{table}[t]
\caption{Termination statistics on Qwen3.6-35B-A3B (3 samples per problem; generation budget of 100K tokens for LiveCodeBench v6 and 70K for GPQA-Diamond, flexible extract). Term.: responses that emit \texttt{</think>} within the budget. Acc.\,$|$\,Term.: accuracy among these responses. Loops: responses that never emit \texttt{</think>} and end in verbatim repetition. All values in \%.}
\label{tab:termination}
\vspace{-0.2in}
\begin{center}
\small
% \begin{tabular}{lccccccccccc}
\begin{tabularx}{\textwidth}{l*{12}{>{\centering\arraybackslash}X}}
\toprule
& & \multicolumn{4}{c}{LiveCodeBench v6} & & \multicolumn{4}{c}{GPQA-Diamond} \\
\cmidrule{3-6} \cmidrule{8-11}
Method & Bits & Acc. & Term. & Acc.\,$|$\,Term. & Loops & & Acc. & Term. & Acc.\,$|$\,Term. & Loops \\
\midrule
BF16      & 16    & 77.6 & 97.2 & 79.8 & 1.5  & & 83.0 & 98.7 & 83.4 & 0.7  \\
GPTQ      & 2.250 & 50.1 & 75.3 & 64.2 & 23.2 & & 72.9 & 87.5 & 80.2 & 12.5 \\
MC-MoE    & 2.358 & 72.3 & 92.6 & 76.1 & 6.9  & & 72.6 & 94.3 & 76.1 & 5.6  \\
GSQ       & 2.125 & 66.4 & 84.0 & 77.0 & 14.0 & & 80.6 & 96.8 & 82.3 & 3.0  \\
\ouralg{} & 2.022 & 70.2 & 90.1 & 77.1 & 9.4  & & 79.3 & 95.1 & 81.9 & 4.7  \\
\bottomrule
\end{tabularx}
\end{center}
\end{table}

\Cref{tab:termination} tests the premise of \ouralg{} directly. When quantized models close their reasoning, they answer nearly as accurately as BF16: for GSQ and \ouralg{}, accuracy among terminated responses stays within 3 points of BF16 on both benchmarks. Much of the accuracy loss instead comes from responses that never terminate, most of which end in verbatim repetition loops. Among the questions that BF16 answers correctly but a quantized baseline does not, 19--71\% of the failures are non-terminated responses across LiveCodeBench, GPQA-Diamond and AIME 2026. Forcing termination confirms this: truncating the reasoning of a quantized model at the BF16 reasoning length for the same question, appending \texttt{</think>}, and letting the same model answer recovers 41\% (GPTQ) and 55\% (MC-MoE) of the non-terminated failures on LiveCodeBench.
Termination failure is less pronounced on Nemotron-3-Nano-30B-A3B, where truncation stays at 1--10\% for every method and the gap comes mostly from terminated responses; \ouralg{} still attains the highest accuracy among the quantized models on both benchmarks.

\section{Choice of Protected Experts}
\label{sec:abl-selection}

\begin{table}[t]
\caption{Choice of the 223 experts held at 3 bits on Qwen3.6-35B-A3B (LiveCodeBench v6), all under the same budget of 2.022 bits. None: the 223 experts are allocated by the knapsack like the others. Paired: number of responses that terminate only with $\mathcal{P}$ versus only with the alternative set.}
\label{tab:selection}
\vspace{-0.2in}
\begin{center}
\small
% \begin{tabular}{lccccc}
\begin{tabularx}{\textwidth}{l*{6}{>{\centering\arraybackslash}X}}
\toprule
Protected experts & Acc. (\%) & Term. (\%) & Loops (\%) & Length & Paired \\
\midrule
None                      & 69.7 & 85.2 & 14.0 & 42.1K & -- \\
$\mathcal{P}$ (ours)      & \textbf{70.2} & \textbf{90.1} & \textbf{9.4} & \textbf{37.0K} & -- \\
Random                    & 68.4 & 81.4 & 17.8 & 43.9K & 55\,:\,21 \\
Top by routing frequency  & 68.7 & 86.3 & 13.0 & 41.3K & 36\,:\,21 \\
Top by Fisher sensitivity & 66.4 & 82.2 & 16.0 & 44.1K & 51\,:\,20 \\
\bottomrule
\end{tabularx}
\end{center}
\end{table}

To test whether the gain comes from \emph{which} experts are protected rather than from the extra bits, we hold four different sets of 223 experts at 3 bits under the same budget (\cref{tab:selection}). Only $\mathcal{P}$ improves termination over the unprotected model (85.2\% $\rightarrow$ 90.1\%) and reduces loops (14.0\% $\rightarrow$ 9.4\%). Random, routing-frequency, and Fisher-selected sets stay near or below the unprotected model (81.4--86.3\% termination) and produce longer outputs (41.3K--44.1K tokens vs.\ 37.0K). The termination gain of $\mathcal{P}$ is significant against the random and Fisher-selected sets (exact McNemar test, $p < 10^{-3}$) but not against the frequency-selected set ($p = 0.06$).
The same specificity appears in the BF16 model: blocking routing to $\mathcal{P}$ lowers the termination rate on GPQA-Diamond from 98.0\% to 60\%, whereas blocking two random sets of 223 experts with matching routing frequency leaves it at 98.0--98.5\%.

\begin{table}[t]
\caption{Protecting $\mathcal{P}$ at two bit budgets on Qwen3.6-35B-A3B (LiveCodeBench v6). Acc.\,$|$\,Term.: accuracy among responses that emit \texttt{</think>}.}
\label{tab:low-bit}
\vspace{-0.2in}
\begin{center}
\small
% \begin{tabular}{llccccc}
\begin{tabularx}{\textwidth}{l*{7}{>{\centering\arraybackslash}X}}
\toprule
Bits & $\mathcal{P}$ at 3 bits & Acc. (\%) & Term. (\%) & Acc.\,$|$\,Term. (\%) & Loops (\%) & Length \\
\midrule
2.02 & \checkmark & 70.2 & 90.1 & 77.1 & 9.4  & 37.0K \\
2.02 & $\times$   & 69.7 & 85.2 & 77.6 & 14.0 & 42.1K \\
\midrule
1.63 & \checkmark & 34.6 & 37.4 & 78.2 & 61.8 & 71.3K \\
1.63 & $\times$   & 22.7 & 28.2 & 61.3 & 71.5 & 77.1K \\
\bottomrule
\end{tabularx}
\end{center}
\end{table}

\paragraph{Lower bit budgets.}
The benefit of protecting $\mathcal{P}$ grows as the budget shrinks (\cref{tab:low-bit}). At 1.63 bits, holding $\mathcal{P}$ at 3 bits raises accuracy from 22.7\% to 34.6\% and termination from 28.2\% to 37.4\%, compared with a gain of 0.5 points at 2.02 bits. With $\mathcal{P}$ protected, accuracy among terminated responses stays at 78.2\%, close to the 2.02-bit level, whereas without protection it falls to 61.3\%: at this budget, the experts that decide when to stop are themselves heavily quantized.

\section{Sensitivity to $\tau$ and Stability of $\mathcal{P}$}
\label{sec:abl-tau}

\begin{table}[t]
\caption{Termination-aware experts re-identified on seven datasets disjoint from all evaluation benchmarks (Qwen3.6-35B-A3B, BF16 trajectories). Selected and In $\mathcal{P}$ use $\tau{=}0.8$. The last four columns give the fraction of the selected experts that belong to $\mathcal{P}$ for each threshold $\tau$.}
\label{tab:p-stability}
\vspace{-0.2in}
\begin{center}
\small
\setlength{\tabcolsep}{5pt}
% \begin{tabular}{lrrrcccc}
\begin{tabularx}{\textwidth}{l*{8}{>{\centering\arraybackslash}X}}
\toprule
Dataset & Positions & Selected & In $\mathcal{P}$ & $\tau{=}0.6$ & $0.7$ & $0.8$ & $0.9$ \\
\midrule
APPS             & 71  & 181 & 160 & 0.75 & 0.79 & 0.88 & 0.99 \\
CodeContests     & 51  & 177 & 157 & 0.74 & 0.81 & 0.89 & 0.93 \\
MATH             & 89  & 163 & 157 & 0.80 & 0.87 & 0.96 & 1.00 \\
NuminaMath       & 87  & 164 & 162 & 0.83 & 0.92 & 0.99 & 1.00 \\
ARC              & 95  & 177 & 164 & 0.81 & 0.87 & 0.93 & 0.98 \\
OpenBookQA       & 96  & 214 & 174 & 0.72 & 0.76 & 0.81 & 0.87 \\
Nemotron-Science & 364 & 149 & 146 & 0.85 & 0.93 & 0.98 & 1.00 \\
\bottomrule
\end{tabularx}
\end{center}
\end{table}

We re-identify termination-aware experts on seven datasets that are disjoint from all evaluation benchmarks, covering code (APPS, CodeContests), math (MATH, NuminaMath), and science (ARC, OpenBookQA, Nemotron-Science). On every dataset, 81--99\% of the selected experts belong to $\mathcal{P}$ (\cref{tab:p-stability}), so $\mathcal{P}$ reflects the model rather than the data used to identify it. Raising $\tau$ pushes this fraction toward 1.0, which indicates that $\tau$ controls how many experts are selected rather than which ones.
Accuracy is also insensitive to the size and precision of the protected set. Lowering $\tau$ enlarges $\mathcal{P}$ from 223 to 336 experts and gives 69.0\% on LiveCodeBench (70.2\% at $\tau{=}0.8$), and protecting a larger set of 471 experts at 3 or 4 bits gives 71.3\% and 70.5\%. We therefore use $\tau{=}0.8$ with 3-bit protection.

\section{Entire Quantization Algorithm}\label{sec:entire_algorithm}
Our pipeline consists of three stages:
(i) quantization-bank construction,
(ii) termination-aware mixed-precision allocation, and
(iii) codebook-constrained error correction.
Only the routed-expert weights are quantized; all remaining model
components stay in BF16.
Algorithms~\ref{alg:bank}--\ref{alg:correction} summarize the full
procedure.
\begin{algorithm}[t]
\caption{Stage 1: Quantization-Bank Construction}
\label{alg:bank}
\begin{algorithmic}[1]

\Require BF16 routed-expert matrices and routed calibration inputs
\Ensure Bank
$\mathcal{B}=\{q_K,c_K,\mu_K,\alpha_K\}$
for all candidate $K$

\For{each expert matrix $W \in \{W_{gu}^{e},W_d^{e}\}$}

    \State Compute the routed-input second moment
    $S \gets \sum_t x_t x_t^{\top}$

    \State Row-normalize $W$ using
    \cref{eq:normalization}

    \State Construct the activation-weighted histogram using
    $m_{ij}=(\alpha_i^{(0)})^2 S_{jj}$

    \State Run one exact 1-D dynamic program up to $K_{\max}$

    \State Recover and gauge-fix the codebook $c_K$
    for every candidate $K$

    \For{each candidate $K$}

        \If{$K=3$}
            \State Obtain ternary codes using
            Equations~\ref{eq:ternary_codes}--\ref{eq:ternary_offset}
        \Else
            \State Refine $(q,\mu,\alpha)$ by Step-2 coordinate
            descent
        \EndIf

        \State Refit $(\mu,\alpha)$ using the full $S$ via
        \cref{eq:stage3_normal_equations}

        \State Store $\{q_K,c_K,\mu_K,\alpha_K\}$ in $\mathcal{B}$

    \EndFor

\EndFor

\State \Return $\mathcal{B}$

\end{algorithmic}
\end{algorithm}

\begin{algorithm}[!htbp]
\caption{Stage 2: Termination-Aware Mixed-Precision Allocation}
\label{alg:allocation}
\begin{algorithmic}[t]

\Require Quantization bank $\mathcal{B}$, calibration documents,
routing traces, target bitwidth $\bar{b}$, pin level $K_{\mathrm{pin}}$
\Ensure Allocated level $K_{l,e}$ for every routed expert

\State Collect BF16 block-output gradients
$g_t=\partial\mathcal{L}_r/\partial h_t$

\For{each layer $l$, expert $e$, and candidate level $K$}

    \State Decode the corresponding bank variant

    \State Compute
    $\delta_t =
    w_{t,e}(f_e^K(x_t)-f_e(x_t))$

    \State Compute Fisher sensitivity
    $C[l,e,K]$ using
    \cref{eq:fisher_expert_cost}

    \State Compute the optional Fisher-free output cost
    $C_{\mathrm{out}}[l,e,K]$

\EndFor

\State Identify the termination-expert set
$\mathcal{P}$ using
\cref{eq:termination_aware}

\State Pin
$K_{l,e}\gets K_{\mathrm{pin}}$
for all $(l,e)\in\mathcal{P}$

\State Restrict all remaining experts to the trust region
$\mathcal{K}_{l,e}$

\State Solve the multiple-choice knapsack by Lagrangian bisection

\State Fill any remaining discrete budget greedily by the largest
cost reduction per added bit

\State Use the Fisher-cost allocation; the held-out calibration NLL is used only to choose the trust region $\mathcal{K}_{l,e}$

\State \Return $\{K_{l,e}\}$

\end{algorithmic}
\end{algorithm}

\begin{algorithm}[t]
\caption{Stage 3: Codebook-Constrained Error Correction}
\label{alg:correction}
\begin{algorithmic}[t]

\Require Bank $\mathcal{B}$ and allocated level $K_{l,e}$ for each expert
\Ensure Final packed mixed-precision routed experts

\For{each routed expert $e$ at its allocated level $K$}

    \State Construct and damp
    $H_{gu}^{e}
    =\sum_t w_{t,e}^{2}x_t x_t^{\top}$

    \State Reassign $W_{gu}^{e}$ codes within the fixed codebook
    using GPTQ/OBS feedback

    \State Refit $(\mu,\alpha)_{gu}$ using
    \cref{eq:final_affine_refit}

    \State Compute corrected intermediate activations
    $\widetilde{h}_t$

    \State Construct and damp
    $H_d^{e}
    =\sum_t
    w_{t,e}^{2}
    \widetilde{h}_t\widetilde{h}_t^{\top}$

    \State Reassign $W_d^{e}$ codes within the fixed codebook

    \State Refit $(\mu,\alpha)_d$

    \State Fit the scalar output gain
    $\gamma_e$ using
    \cref{eq:scalar_gain}

    \State Fold $\gamma_e$ into the down-projection scale

    \State Verify that all reconstructed weights remain on the
    allocated codebook grid

\EndFor

\State \Return final quantized routed experts

\end{algorithmic}
\end{algorithm}

\subsection{Guarantees}
\label{sec:guarantees}

The pipeline combines exact sub-problems with local approximations and
greedy procedures.
Table~\ref{tab:guarantees} summarizes which statements are guaranteed
by the corresponding optimization problem and which depend on
approximations.

\begin{table}[t]
\caption{Exact and approximate components of the proposed pipeline.}
\label{tab:guarantees}
\begin{center}
\small
\setlength{\tabcolsep}{4pt}
\renewcommand{\arraystretch}{1.15}

\begin{tabular}{
    p{0.22\textwidth}
    p{0.51\textwidth}
    p{0.20\textwidth}
}
\hline
{\bf STEP}
&
{\bf STATEMENT}
&
{\bf STATUS}
\\
\hline \\[-1.2ex]

Codebook
&
For every $K$, the dynamic program returns the global minimum of the
weighted one-dimensional $k$-means objective defined on the histogram.
&
Exact
(Lemma~1,
Proposition~2)
\\[1.0ex]

Step-2 codes
&
Alternating code assignment and affine fitting never increases the
unweighted row reconstruction error and terminates at a fixed point.
&
Monotone descent
\\[1.0ex]

Affine relocation / refit
&
For fixed codes, the $S$- or $H$-weighted row reconstruction error is
minimized over $(\mu,\alpha)$.
&
Exact convex
$2\times2$
solve
\\[1.0ex]

Sensitivity cost
&
\cref{eq:fisher_expert_cost} estimates the loss increase using
a BF16-anchored second-order expansion and diagonal empirical-Fisher
curvature.
&
Approximate
(A4--A6)
\\[1.0ex]

Bit allocation
&
For a fixed $\lambda$, the Lagrangian allocation is optimal among
allocations whose average bitwidth does not exceed the discrete
bitwidth that it actually attains.
Remaining budget slack is filled greedily.
&
Proposition~1
$+$ greedy slack
\\[1.0ex]

Code reassignment
&
At each processed column, OBS gives the minimum-curvature compensation
subject to the current rounding decision; the full sweep is greedy over
the column order and preserves the allocated codebook format.
&
Exact per step;
greedy globally
\\[1.0ex]

Output gain
&
For fixed corrected expert outputs, $\gamma_e$ minimizes the summed
squared output mismatch with the BF16 expert.
&
Exact scalar
least squares
\\[1.0ex]

\hline
\end{tabular}
\end{center}
\end{table}

\subsection{Bits per Weight}\label{app:reuse_storage}
\emph{Non-uniform} quantization learns the levels, typically using $k$-means
over the weight values \citep{han2016deep, kim2024}, thereby allocating more
levels to regions with greater probability mass; ternary quantization
\citep{li2016ternary, ma2024bitnet} is the case $K=3$ with $c=(-1,0,+1)$ and a
per-row scale. The codes require $\log_2 K$ bits per weight, and the two affine
parameters, each stored in BF16, contribute a further $32/p$. Ignoring the
negligible codebook overhead, the effective bitwidth of
\cref{eq:quant_representation} is therefore
\begin{equation}
    b_{\mathrm{eff}}
    =
    \log_2 K + \frac{32}{p}.
    \label{eq:effective_bitwidth}
\end{equation}
For an expert matrix using a $K$-level codebook, the discrete code
requires a logical $b_K = \log_2 K$ bits per weight.
For non-power-of-two $K$, we report this logical precision when
describing the mixed-precision allocation.
The physical packing can differ slightly; for example, five ternary
weights can be packed into one byte, corresponding to $1.6$ physical
bits per weight.

Each matrix also stores one BF16 offset and one BF16 scale for every
output row.
For a matrix with $p$ input columns, this contributes
$b_{\mathrm{affine}} = 32/p$ bits per weight:
$32/2048 = 0.015625$ for the fused gate/up projection ($p=2048$) and
$32/512 = 0.0625$ for the down projection ($p=512$).
Because an expert contains twice as many weights in $W_{gu}$ as in
$W_d$, the parameter-weighted affine overhead over a complete expert is
$b_{\mathrm{affine}}^{\mathrm{expert}} = (2 \cdot 0.015625 + 0.0625)/3 = 0.03125$
bits per routed-expert weight.

The per-matrix codebook contains only $K$ BF16 scalars and is
negligible relative to the matrix itself.
The expert gain $\gamma_e$ is folded into the down-projection scale and
therefore adds no separate runtime or storage term.

Unless explicitly stated otherwise, the reported
\emph{average bitwidth} refers to the logical code precision
\begin{equation}
    \bar{b}
    =
    \frac{1}{N}
    \sum_{l,e}
    \log_2 K_{l,e},
    \label{eq:reported_average_bits}
\end{equation}
averaged over the $N$ equally sized routed experts.
Including the row-affine overhead, a reported $2.000$-bit allocation
occupies approximately $2.000 + 0.03125 = 2.03125$ bits per
routed-expert weight.
Similarly, for the outside-budget pinned allocation in
\cref{eq:pinned_bitwidth}, whose logical average precision is
$\bar{b}'=2.0218$, the corresponding storage precision is
approximately $2.0218 + 0.03125 = 2.05305$ bits per routed-expert
weight.

\section{Acknowledgment}
We thank Yelim Lee for her assistance in preparing the figures for this paper.

\end{document}